\pdfoutput=1

\documentclass[11pt]{article}

\usepackage[final]{acl}

\usepackage{times}
\usepackage{latexsym}

\usepackage[T1]{fontenc}

\usepackage[utf8]{inputenc}

\usepackage{microtype}

\usepackage{inconsolata}

\usepackage{graphicx}
\usepackage{subcaption}

\usepackage{algorithm}
\usepackage{multirow}
\usepackage{graphicx}
\usepackage{booktabs}
\usepackage{algorithm}
\usepackage{multicol}
\usepackage{amsmath}
\usepackage{algpseudocode}
\usepackage{caption}
\usepackage{enumitem}
\usepackage{microtype}
\usepackage{hyperref}
\usepackage{url}
\usepackage{booktabs}

\usepackage{tabularray}
\usepackage{multirow}
\usepackage{subfiles}
\usepackage{enumitem}
\usepackage{makecell}
\usepackage{graphicx}
\usepackage{wrapfig}
\definecolor{mydarkblue}{rgb}{0,0.08,0.45}
\definecolor{myblue}{HTML}{3b75c9}
\definecolor{myred}{HTML}{E33222}
\definecolor{mygreen}{HTML}{438773}
\definecolor{mymaroon}{RGB}{142,27,19}
\definecolor{maroon}{HTML}{992000}
\definecolor{mycite}{cmyk}{0.55,1,0,0.15}
\definecolor{codeblue}{rgb}{0.25,0.5,0.5}
\definecolor{codekw}{rgb}{0.85, 0.18, 0.50}
\definecolor{codegreen}{rgb}{0,0.6,0}
\definecolor{codegray}{rgb}{0.5,0.5,0.5}
\definecolor{codepurple}{rgb}{0.58,0,0.82}
\definecolor{backcolour}{rgb}{0.95,0.95,0.92}
\definecolor{refcolor}{HTML}{9F363A}
\hypersetup{
    colorlinks=true,
    citecolor=refcolor,
    linkcolor=myblue
          }

\title{Pretrained Image-Text Models are Secretly Video Captioners}

\author{
    Chunhui Zhang$^{*}$ \quad Yiren Jian$^{*}$ \quad Zhongyu Ouyang \quad Soroush Vosoughi\\
    Department of Computer Science, Dartmouth College \\
    \small{\texttt{\{chunhui.zhang.gr, yiren.jian.gr, zhongyu.ouyang.gr, soroush.vosoughi\}@dartmouth.edu}}
}

\begin{document}
\newcommand\blfootnote[1]{%
  \begingroup
  \renewcommand\thefootnote{}\footnote{#1}%
  \addtocounter{footnote}{-1}%
  \endgroup
}

\maketitle
\blfootnote{$^{*}$Equal contribution and random order.}

\begin{abstract}
Developing video captioning models is computationally expensive. The dynamic nature of video also complicates the design of multimodal models that can effectively caption these sequences. However, we find that by using minimal computational resources and without complex modifications to address video dynamics, an image-based model can be repurposed to outperform several specialised video captioning systems. Our adapted model demonstrates top-tier performance on major benchmarks, ranking 2nd on MSR-VTT and MSVD, and 3rd on VATEX. We transform it into a competitive video captioner by post-training a typical image captioning model BLIP-2 with \textit{only} 6,000 video-text pairs and \textit{simply} concatenating frames—significantly fewer data than other methods, which use 2.5 to 144 million pairs. From a resource optimization perspective, this video captioning study focuses on three fundamental factors: optimizing model scale, maximizing data efficiency, and incorporating reinforcement learning. This extensive study demonstrates that a lightweight, image-based adaptation strategy can rival state-of-the-art video captioning systems, offering a practical solution for low-resource scenarios.
\end{abstract}

\section{Introduction}
\label{sec:intro}
Vision-language pretraining significantly advances multimodal tasks such as captioning, question answering, retrieval and broader video understanding~\citep{liu2023llava, liu2023improvedllava, li2023blip, dai2023instructblip, chen2023vast, kuo2023mammut, mplug2, diao2023av, diao2024learning, diao2025temporal, zhang2022look, liu-etal-2024-infimm, han2024infimmwebmathb, jian2023bootstrapping, jian2024expedited}. Among these, video captioning stands out as it narrates visual concepts and their temporal interactions, reflecting the intricate multimodal processes as humans to perceive and articulate dynamic visual experiences.

Current video-text methods often incorporate intricate designs tailored to video inputs. For instance, some models extend existing frameworks by integrating frame samplers to capture temporal dynamics~\citep{alayrac2022flamingo, yang2021, xu2021videoclip}. Other approaches, such as ALPRO~\citep{li2021alignprompt} and VIOLET~\citep{fu2023empiricalmvm}, propose end-to-end models that are meticulously trained on large-scale video-text datasets sourced from the Web~\citep{zellers2021merlot, Bain21}.
Despite their success, video captioning models remain highly resource-intensive, often hitting performance bottlenecks when \textit{(i)} computational resources are constrained, or \textit{(ii)} the task requires specialized priors without clear guidance for model design and training.
This raises a critical question: \textbf{for simplicity and efficiency, how can we repurpose existing image captioning models for video captioning, without relying on complex, hand-crafted video-specific designs?}

\begin{figure*}[t]
\centering
\includegraphics[width=1\textwidth]{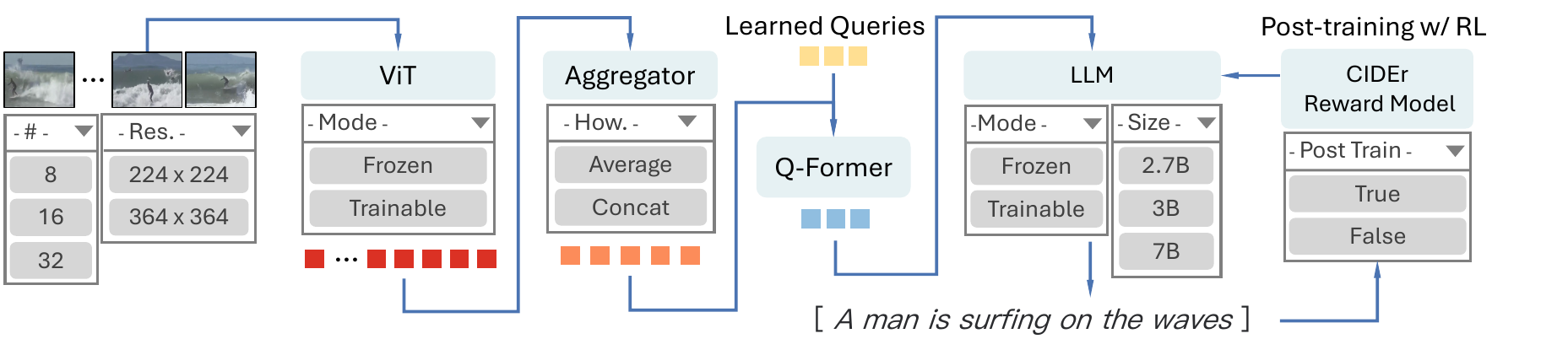}
\vspace{-0.25in}
\caption{Key factors in recycling BLIP for video captioning: \textbf{Model} – assessing the scale and trainability of components like the ViT, LLM, and Q-Former; \textbf{Data} – examining the volume, quality, and fusion strategies for image and video-text pairs; \textbf{Supervision} – employing reinforcement learning to align generated captions with human language quality standards (CIDEr).}
\vspace{-0.05in}
\label{fig:pipeline}
\end{figure*}
To address this, we revisit fundamental factors in training—\textbf{model scale}, \textbf{data efficiency}, and \textbf{supervision}—that critically influence video captioning while being agnostic to the variants of video-specific designs:
\underline{First}, we find that moderate-sized language models (LMs) when fine-tuned for specific tasks, can meet the demands of video captioning efficiently. This challenges the common belief that larger models are always superior, demonstrating that targeted optimization can outperform sheer model size.
\underline{{Second}}, using extensive pretraining on image-text pairs, as demonstrated with BLIP-2, is transferable to video tasks. This allows the model to achieve high performance with minimal video usage, offering an efficient alternative to training from scratch.
\underline{Third}, instead of relying on traditional cross-entropy loss, we optimize directly for non-differentiable CIDEr with reinforcement learning, ensuring that the generated captions better align with human-standard video descriptions.

By bypassing complex, specialized video input designs, our experiments demonstrate that BLIP-2 straightforwardly derived from image captioning, can be effectively optimized to deliver competitive video captioning performance. This study underscores the potential of simplicity and efficiency in advancing multimodal video captioning, providing a streamlined yet stable solution. 
The codes are released: \url{https://github.com/chunhuizng/mllm-video-captioner}.

\section{Recycling BLIP-2 for Video Captioning}
\label{sec:msvtt}
As shown in Fig.~\ref{fig:pipeline}, we adapt BLIP-2, a typical image-text model (details in App.~\ref{app:sec:preliminary}), for video captioning without any additional parameters. Each video frame is encoded by ViT, which generates visual tokens that are concatenated to form a unified representation (e.g., an 8-frame video produces a token sequence of size 8$\times$256). This unified token sequence is then processed by the Q-former and passed to the LM to generate captions.

\section{Training Recipes: Model, Data, and Supervision} 
According to Tab.~\ref{tab:result}, our solution has top-level performance on important benchmarks (particularly on the CIDEr metric-the primary ranking measure on \texttt{Paperswithcode}), ranking 2nd on MSR-VTT and MSVD, and 3rd on VATEX, among models with publicly available code. More importantly, it proves to be highly efficient without any video architecture design, using only \textbf{6k} video-text pairs---significantly less than the \textbf{million-level} datasets required by competing baselines. 

Additional background is in App.~\ref{app:sec:related-work}. The settings are detailed in App.~\ref{app:exp}, and further experiments (\textbf{ablations, {other datasets}, and other video tasks}) supporting the following analysis are in App.~\ref{app:ana}.
\begin{table*}
\centering
\resizebox{1\textwidth}{!}{
\begin{tblr}{
cells = {l},
  cell{1}{1} = {r=2}{},
  cell{1}{2} = {c=5}{},
  cell{1}{7} = {c=5}{},
  cell{1}{12} = {c=5}{},
  hline{1,13} = {-}{0.08em},
  hline{2} = {2-5,6-10,11-15,16-18}{0.03em},
  hline{3, 11} = {-}{0.02em},
  cell{1}{17} = {r=2}{c}, 
  cell{1}{18} = {r=2}{c}, 
  hline{Z} = {-}{0.08em},
}
Model     & MSR-VTT~\citep{xu2016msr} &        &         &        &        & MSVD~\citep{chen2011collecting} &        &         &        &        & VATEX~\citep{wang2019vatex} &        &         &      &     &Code&{\# msr v.-\\t. pairs}\\
          & C.   & M. & R. & B4. & P. & C.   & M. & R. & B4. & P. & C.   & M. & R. & B4. & P. & &-\\
IcoCap    & 60.2    & 31.1   & 64.9    & 47.0   & -   & 110.3 & 39.5   & 76.5    & 59.1   & -   & 67.8  & 25.7   & 53.1    & 37.4   & -   &No &-\\
MaMMUT    & 73.6    & -      & -       & -      & 77.5 & 195.6 & -      & -       & -      & 85.6 & -     & -      & -       & -      & 79.9 &No &-\\
VideoCoCa & 73.2    & -      & 68.0    & 53.8   & -   & -     & -      & -       & -      & -   & 77.8  & -      & 54.5    & 39.7   & -   &No &144.7M\\
VALOR     & 74.0    & 32.9   & 68.0    & 54.4   & 81.0 & 178.5 & 51.0   & 87.9    & 80.7   & 83.7 & 95.8  & 29.4   & 57.4    & 45.6   & 73.3 &Yes &1.18M\\
VLAB      & 74.9    & 33.4   & 68.3    & 54.6   & -   & 179.8 & 51.2   & 87.9    & 79.3   & -   & -     & -      & -       & -      & -   &No &10.7M\\
GIT2      & 75.9    & 33.1   & 68.2    & 54.8   & 75.4 & -     & -      & -       & -      & -   & -     & -      & -       & -      & -   &Yes &-\\
VAST      & 78.0    & -      & -       & 56.7   & 77.2 & -     & -      & -       & -      & -   & 99.5  & -      & -       & 45.0   & 81.9 &Yes &27M\\
mPLUG-2   & 80.0    & 34.9   & 70.1    & 57.8   & 82.7 & 165.8 & 48.4   & 85.3    & 70.5   & 82.5 & -     & -      & -       & -      & -   &Yes &2.5M\\
Ours      & 79.5    & 34.2   & 68.3    & 52.4   & 81.2 & 168.0 & 48.3   & 85.8    & 73.5   & 84.4 & 87.1  & 29.1   & 56.7    & 43.3   & 82.1 &Yes &6K\\
\end{tblr}}
\caption{Overall comparison. The results for MSR-VTT, MSVD, and VATEX are from the \texttt{PaperswithCode} open leaderboard. The abbreviations C., M., R., B4., and P. stand for CIDEr, METEOR, ROUGE-L, BLEU-4, and PAC-S~\cite{sarto2023positive}, respectively. We choose CIDEr as the most referential metric, following the \texttt{PaperswithCode}. Tab.~\ref{tab:model-list} has details about configs and references.}
\label{tab:result}
\vspace{-0.05in}
\end{table*}

\subsection{Model Scale}
\paragraph{Trainability: modal connector $>$ LLM $>$ ViT}
\begin{figure*}[!t]
    \centering
    \begin{subfigure}[t]{0.2455\textwidth}
        \centering
        \includegraphics[width=\textwidth]{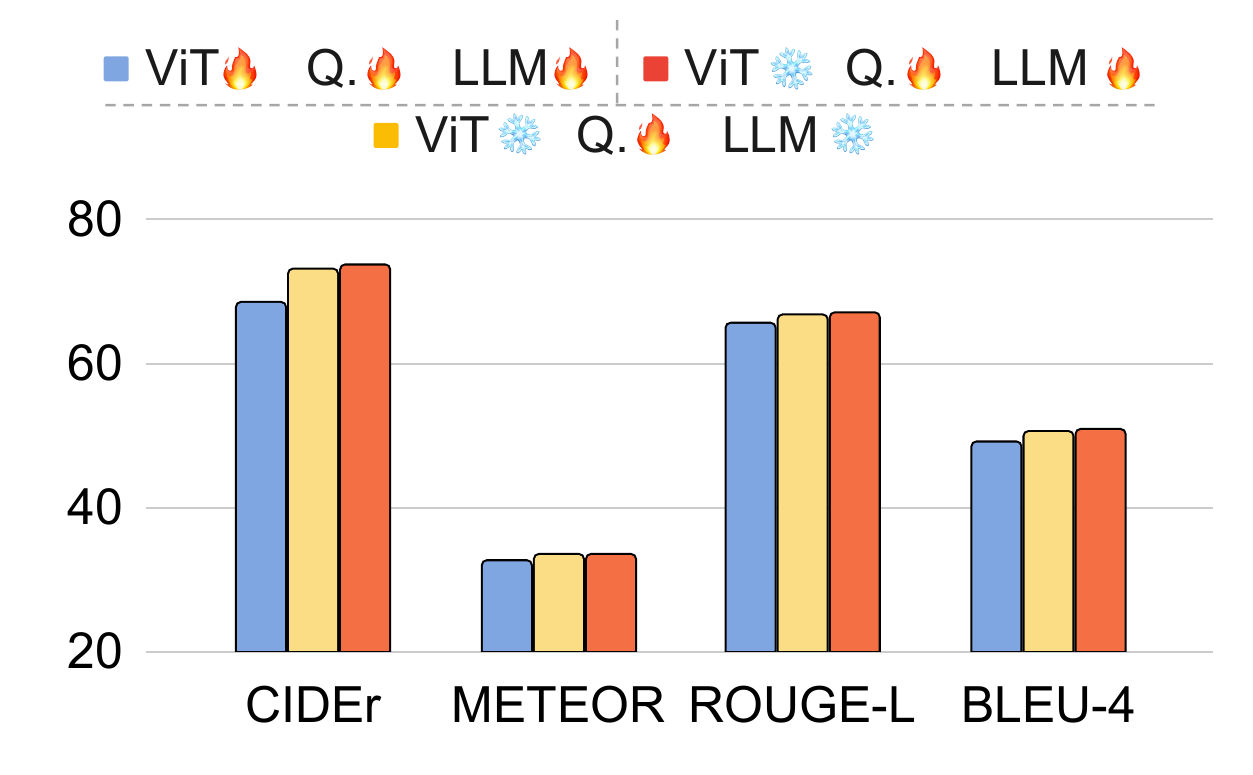}
        \label{fig:trainable-part-combined}
    \end{subfigure}
    \hfill
    \begin{subfigure}[t]{0.2455\textwidth}
        \centering
        \includegraphics[width=\textwidth]{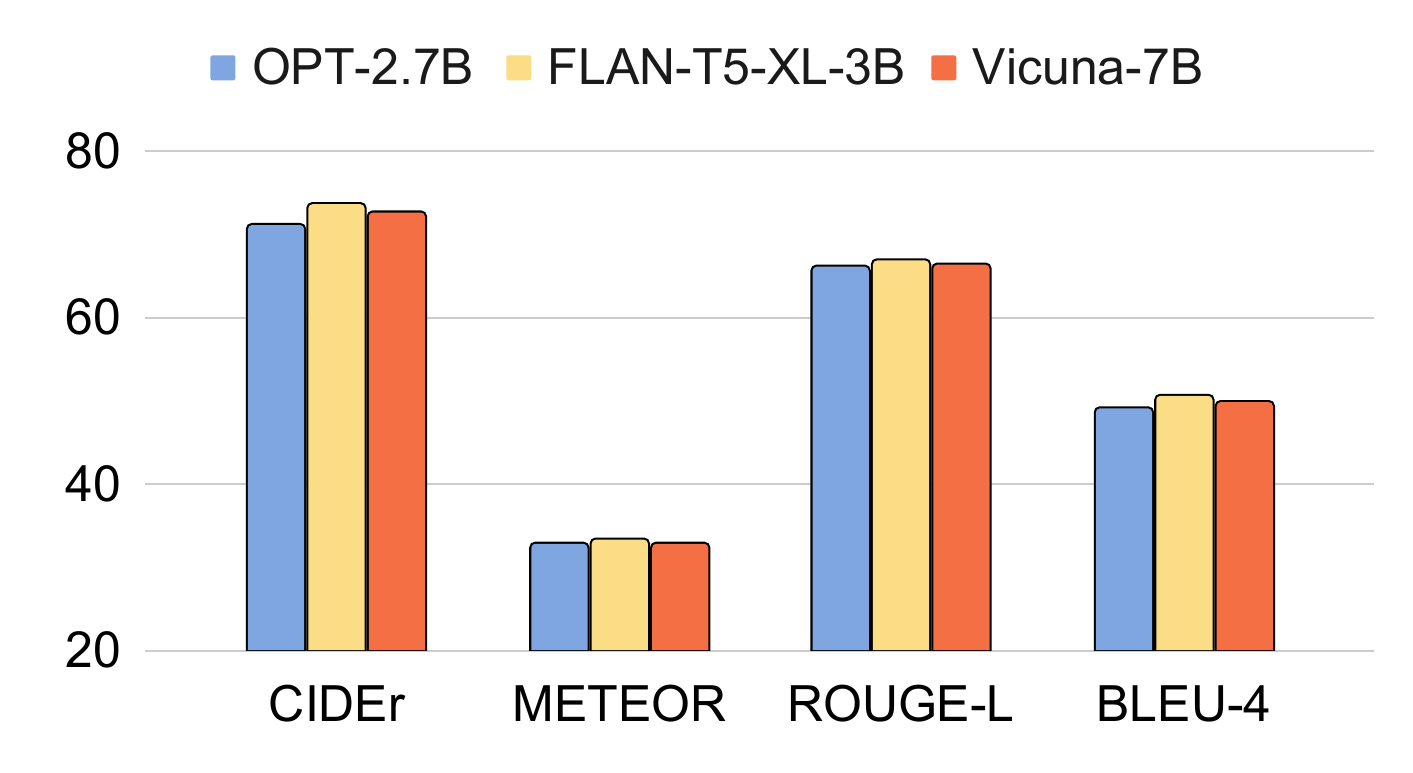}
        \label{fig:small2large-combined}
    \end{subfigure}
        \hfill
    \begin{subfigure}[t]{0.2455\textwidth}
        \centering
        \includegraphics[width=\textwidth]{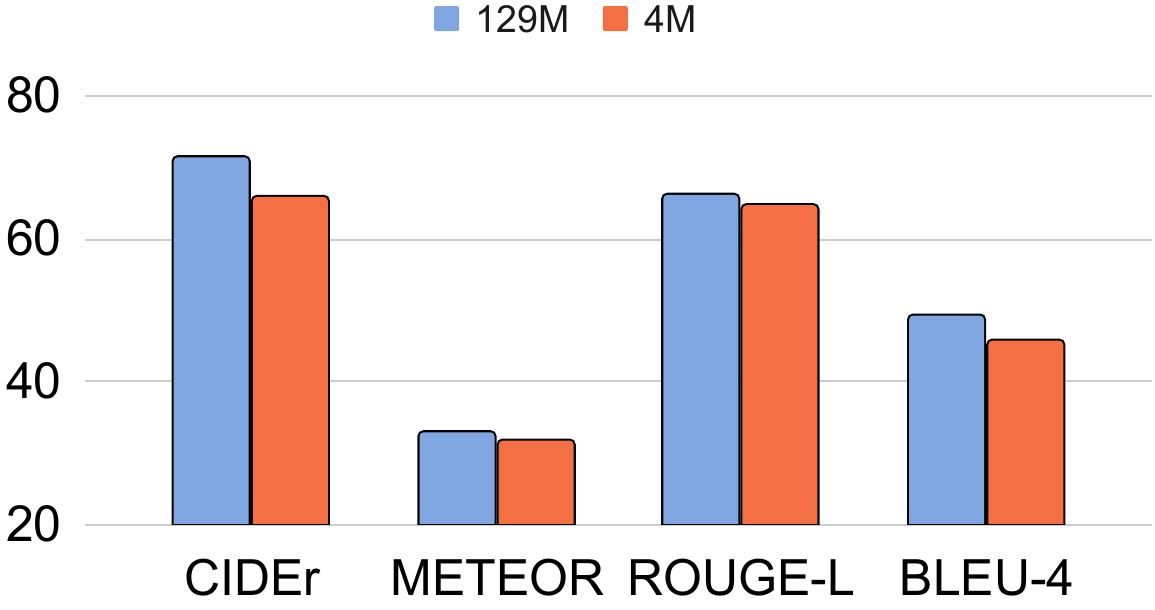}
        \label{fig:pre-train-combined}
    \end{subfigure}
    \hfill
    \begin{subfigure}[t]{0.2455\textwidth}
        \centering
        \includegraphics[width=\textwidth]{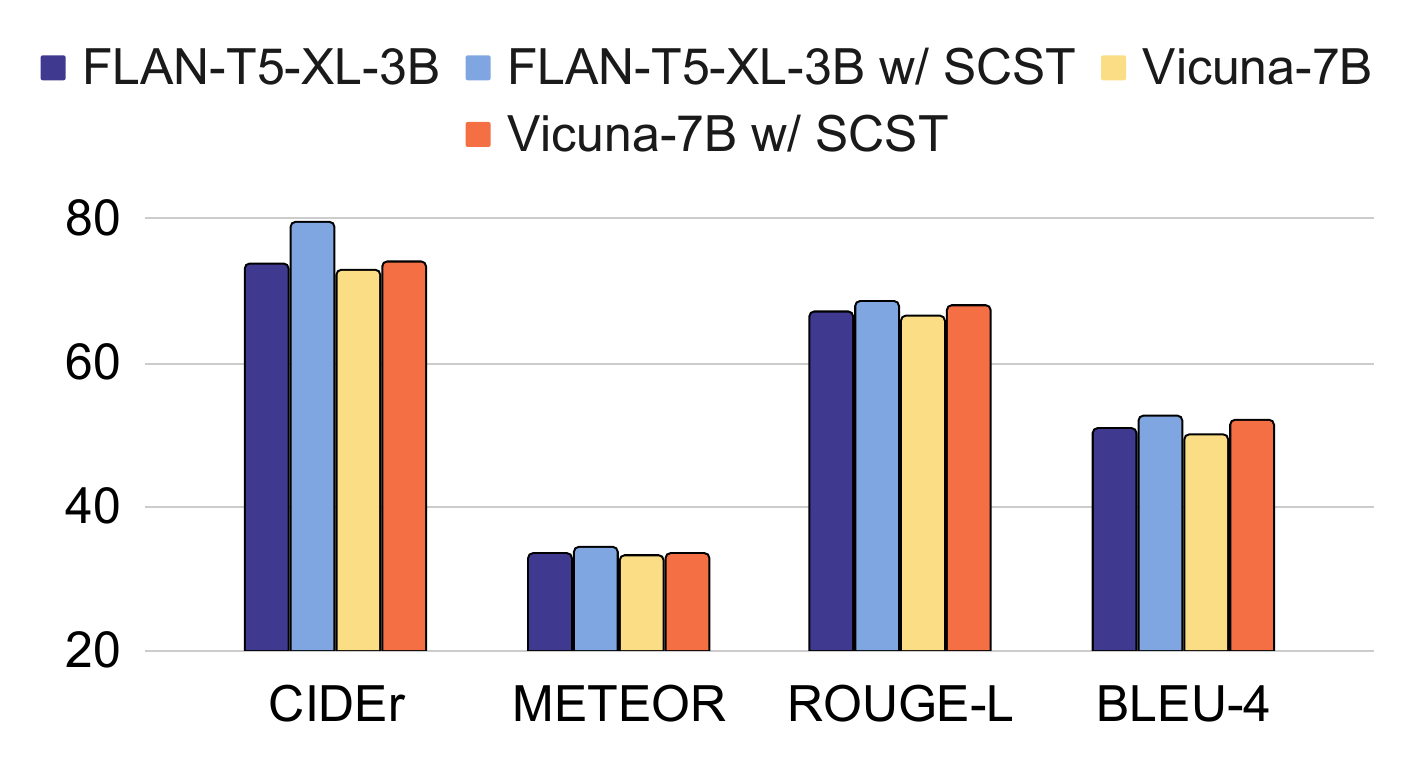}
        \label{fig:scst-bar-combined}
    \end{subfigure}
    \vspace{-0.2in}
    \vspace{-0.1in}
    \caption{Comparisons of different setups for models on the \textit{MSR-VTT} dataset: (a) freezing modules, (b) scales of LLMs, (c) usage of image-text pairs in pretrained BLIP-2, and (d) supervision with and without SCST. We also replicate the comparisons and ablations on other datasets (e.g., \textit{MSVD} and \textbf{VATEX}) in App.~\ref{sec:app:msvd}.}
    \vspace{-0.1in}
    \label{fig:combined}
\end{figure*}
To evaluate the adaptability of various components within the video captioning model, we conducted ablation studies using three setups: training all components, freezing the ViT only, and training the Q-Former only. The results, illustrated in Fig.~\ref{fig:combined}(a) and supported by training curves in Fig.~\ref{fig:app-1} (see App.~\ref{app:sec:qformer-llm-vit} for detailed discussions), reveal a clear performance hierarchy: freezing the ViT (configurations ii and iii) yields higher performance than training all components (configuration i).

Configurations with a frozen ViT allow the Q-Former and LLM to effectively leverage the pre-trained visual features, leading to better alignment in video captioning tasks. Conversely, training the ViT alongside other components introduces potential overfitting and alignment issues, resulting in suboptimal performance. The analysis establishes a hierarchy of trainability: Q-Former $>$ LLM $>$ ViT. The Q-Former shows the highest adaptability during training, followed by the LLM, which benefits from fine-tuning language data. In contrast, the ViT demonstrates the least trainability, as updating its parameters often disrupts the alignment between visual features and language output.

Supporting figures indicate that the Q-Former configuration achieves the most stable performance, reaching peak validation CIDEr scores without significant overfitting (Fig.~\ref{fig:app-1}). This pattern aligns with additional observations in App.~\ref{app:sec:qformer-llm-vit}, confirming that focusing on training the modal connector and LLM while freezing the ViT optimizes the model's performance on video captioning tasks.

\paragraph{Mid-sized LLMs offer trainability for video captioning}
We analyzed the impact of LM size on video captioning by comparing three models: OPT-2.7B, Flan-T5-XL-3B, and Vicuna-7B (see Fig.~\ref{fig:combined}(b) and Fig.~\ref{fig:app:small2large}). 
The BLIP-2 framework was selected for its state-of-the-art performance on the MSCOCO image captioning benchmark, which remains the most canonical dataset for captioning evaluation. The chosen language models—OPT-2.7B, Flan-T5-XL-3B, and Vicuna-7B—are all extensively used within BLIP-2 for vision-language tasks and represent a range of architectures and parameter sizes. Their open-source nature and community adoption further enhance their relevance and comparability in this domain.
The results demonstrate that \textbf{Flan-T5-XL-3B, a mid-sized model, achieves superior performance in generating video captions}, outperforming both the smaller OPT-2.7B and the larger Vicuna-7B on key metric CIDEr. This challenges the notion that larger LMs always yield better results in multimodal tasks.

Training dynamics further support the advantages of mid-sized LLMs. As shown in Fig.~\ref{fig:app:small2large}, the smaller OPT-2.7B model requires 20 epochs to reach peak performance and fails to overfit, indicating limited expressiveness. On the other hand, Vicuna-7B converges rapidly within 5 epochs but quickly shows signs of overfitting, suggesting that its added complexity may not translate into meaningful improvements for video captioning. Flan-T5-XL-3B strikes a balance, reaching peak validation within 14 epochs and maintaining a better trade-off between generalization and overfitting.

These findings and training procedure analysis in App.~\ref{app:sec:mid-size-llm} indicate video captioning tasks{ benefit more from models capable of descriptive processing rather than advanced conversational or reasoning abilities}. Thus, mid-sized LMs like Flan-T5-XL-3B effectively balance trainability, efficiency, and performance in video captioning tasks.

\subsection{Data Efficiency}
\paragraph{Image-Text pretraining offers transferability to video tasks}
We examine the effect of image-text pretraining on video captioning by comparing the performance of two BLIP-2 models pre-trained on different dataset sizes: one on 129 million pairs \textbf{(officially released)} and the other on 4 million pairs \textbf{(reproduced in-house)}. As depicted in Fig.~\ref{fig:combined}(c), the model pre-trained with 129M pairs achieves a significantly higher CIDEr score (71.3) compared to the model trained with only 4M pairs (65.7), underscoring the advantages of using a larger dataset.

Fig.~\ref{fig:app-3} (in App.~\ref{app:sec:image-text-pretrain}) further reveals that the model trained on 129M pairs converges faster and achieves higher performance than the model trained on fewer pairs. This suggests that video captioning tasks require robust grounding, with larger datasets significantly enhancing the model’s ability to map visual concepts to language.

\textbf{These results further underscore the \textit{efficiency} of reusing extensively pre-trained image-text models for video tasks.} Large-scale data exposure improves the model's comprehension of visual content, making it more suitable for generating accurate video captions. For a detailed analysis of the training process, refer to App.~\ref{app:sec:image-text-pretrain}.

\begin{figure}[!t]
    \centering
    \begin{subfigure}[b]{0.225\textwidth}
        \centering
        \includegraphics[width=\textwidth]{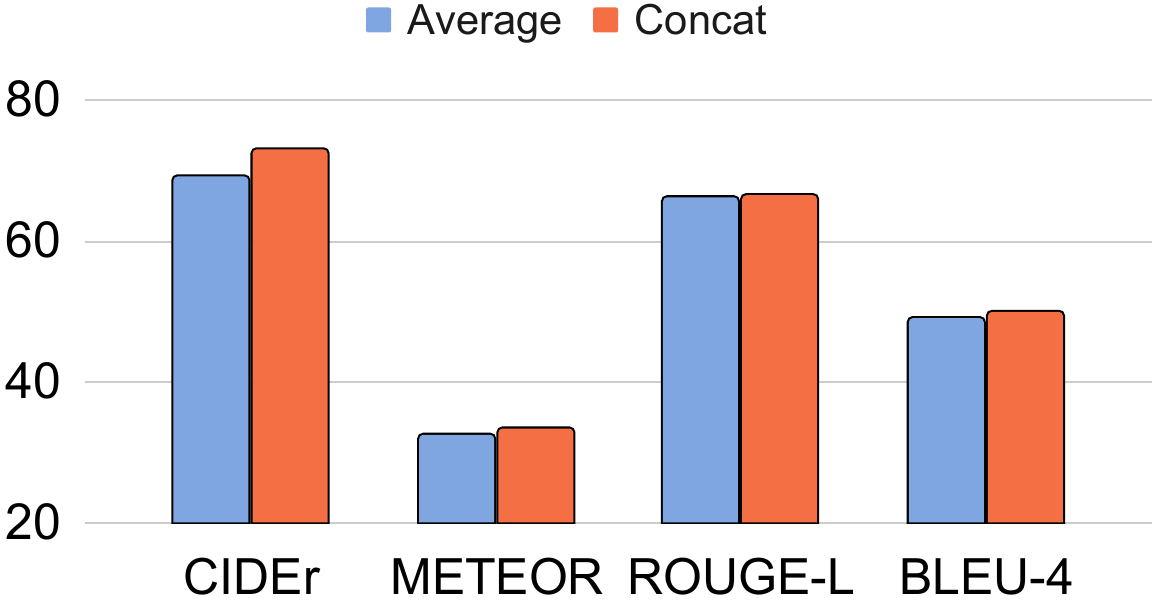} 
    \end{subfigure}
    \hfill
    \begin{subfigure}[b]{0.225\textwidth}
        \centering
        \includegraphics[width=\textwidth]{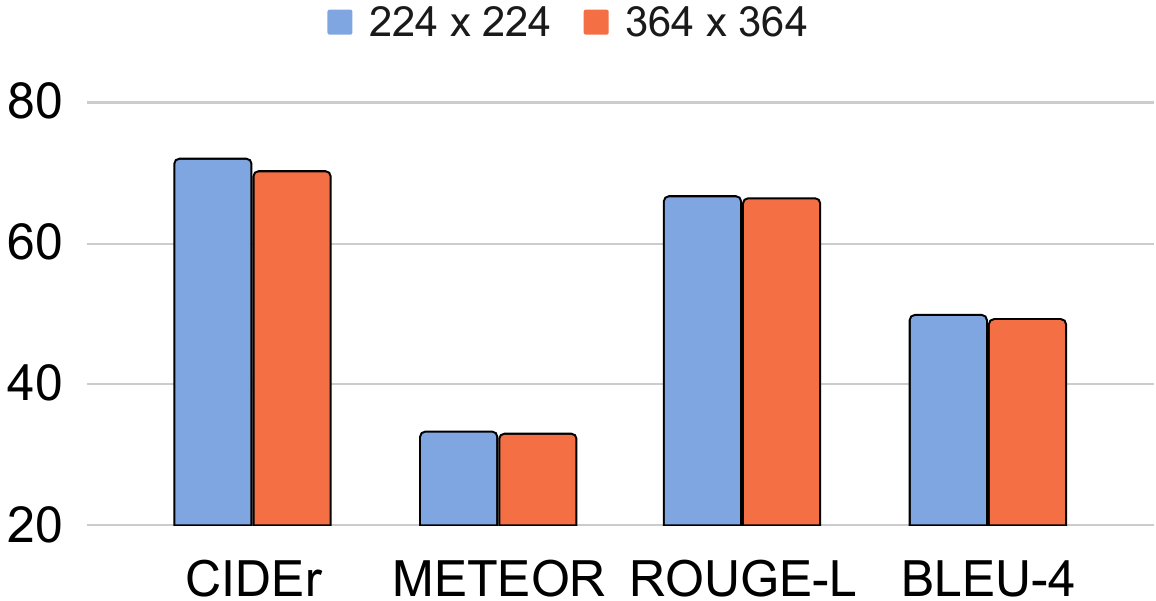} 
    \end{subfigure}
    \caption{(a) temporal fusion by average v.s. concatenation; (b) different resolutions.}
    \vspace{-0.2in}
    \label{fig:combine2}
\end{figure}

\paragraph{Lower resolution efficiently supports video captioning}
We examined the impact of video resolution on training video captioning models by comparing two settings: 224$\times$224 and 364$\times$364. As shown in Fig~\ref{fig:combine2}(b) and~\ref{fig:app-6}, models trained with lower-resolution videos (224$\times$224) achieve competitive performance compared to those trained with higher resolution (364$\times$364), despite exhibiting slightly more fluctuating training curves. 

The results reveal that when basic frame aggregation techniques such as averaging or concatenation are used, lower resolution proves to be not only sufficient but also more efficient for generating accurate captions. The competitive CIDEr obtained with 224$\times$224 resolution indicates that coarse visual information is adequate for the model to perceive and generate descriptive captions effectively. 

Moreover, Fig.~\ref{fig:app-6} demonstrates that while higher resolution (364$\times$364) can lead to more stable training dynamics, the benefits are minimal when sophisticated frame aggregation is not applied. These findings suggest that adopting lower resolution offers practical advantages, including reduced computational requirements, without compromising captioning performance. 
For further insights, see the detailed analysis in App.~\ref{app:sec:lower-resolution}. 
\paragraph{Frame concatenation effectively captures temporality}
We evaluate two approaches for temporal fusion in video captioning: frame averaging and frame concatenation. Frame averaging computes the average of visual tokens across sampled frames, maintaining a fixed dimension. In contrast, frame concatenation extends the token sequence by concatenating visual tokens from each sampled frame, preserving more granular temporal information. These fused tokens are subsequently processed by the Q-Former for caption generation.

The training dynamics, illustrated in Fig.~\ref{fig:app-5} and Fig.~\ref{fig:combine2} (a), show that models using frame concatenation consistently outperform those using frame averaging on CIDEr. The model with frame concatenation reaches peak validation performance around epoch 8 (Fig.~\ref{fig:app-5}), indicating that this method effectively retains temporality. In contrast, frame averaging shows significant performance oscillations after epoch 5, suggesting that it fails to capture sufficient temporal details for stable training.

These findings indicate that frame concatenation is more effective for capturing temporal information in video captioning, as it retains detailed visual context across frames. This approach allows the LM to access a richer set of visual concepts, resulting in more accurate and coherent captions. For additional analysis, see App.~\ref{app-5-frame}.

\subsection{Training Supervision}
\paragraph{Reinforcement learning aligns captioning with human preference}
Traditional video captioning methods often rely on cross-entropy loss, which fails to fully align with human preferences for natural sentence generation. To address this, we use SCST~\citep{rennie2017self}, which directly optimizes toward the human-like CIDEr metric. SCST leverages policy gradients from the non-differentiable CIDEr objective to guide updates to the Q-Former, LLM, and LoRA layers, enhancing alignment with human evaluation standards.

Fig.~\ref{fig:combined}(d) and \ref{fig:app-4} show that SCST improves CIDEr scores by approximately 6.5\% for Flan-T5-XL-3B and 3.4\% for Vicuna-7B, while also boosting other metrics such as METEOR and ROUGE-L. Additionally, Fig.~\ref{fig:app-4} illustrates a decoupling effect between training loss and validation CIDEr; models trained with SCST achieve higher CIDEr scores despite fluctuations in training loss. This shift reflects a prioritization of metrics aligned with human judgment over mere loss minimization.

The smaller improvement for Vicuna-7B likely results from its prior alignment training, which already incorporates reinforcement-based methods. Overall, SCST effectively aligns the training process with human-centered metrics, demonstrating its value for improving video captioning models. See App.~\ref{sec:app:3} for further details.

\section{Discussion and Conclusion}
\label{sec:conclusion}
This study stands out from existing video captioning research by identifying three factors---\textbf{model scale, data efficiency, and training supervision}---that are critical for effectively adapting image captioning models to video tasks. By using these insights to reuse the image-based BLIP-2 model for video tasks, our solution with minimal resource usage ranks \textit{2nd, 2nd, and 3rd} on MSR-VTT, MSVD, and VATEX. This \textbf{open-source guide} provides a foundation for future research aimed at optimizing resource allocation in video captioning and refining post-training techniques.

\section*{Limitations}
Our open-source solution is currently tailored specifically for video captioning tasks due to the page constraints of this short track. While this focus allows for a detailed and resource-efficient guide, it has not shown immediate applicability to other tasks. However, the methods presented can still be extended to broader applications, in particular to facilitate large-scale pseudolabeling for videotext datasets.

This approach is particularly valuable in specialized domains where annotated data is scarce, providing an efficient way to significantly expand video-text data resources. Similar to how the LAION dataset has advanced the image-text field by leveraging BLIP-1 for large-scale pseudolabeling~\citep{li2022blip, schuhmannlaion}, our work aims to bring comparable improvements to video-text integration, enabling further research and development in this area.

\section*{Acknowledgment}
This work was partially funded by a Google Research Award.
\bibliography{ref.bib}

\begin{thebibliography}{53}
\providecommand{\natexlab}[1]{#1}

\bibitem[{Alayrac et~al.(2022)Alayrac, Donahue, Luc, Miech, Barr, Hasson, Lenc, Mensch, Millican, Reynolds et~al.}]{alayrac2022flamingo}
Jean-Baptiste Alayrac, Jeff Donahue, Pauline Luc, Antoine Miech, Iain Barr, Yana Hasson, Karel Lenc, Arthur Mensch, Katherine Millican, Malcolm Reynolds, et~al. 2022.
\newblock Flamingo: a visual language model for few-shot learning.
\newblock In \emph{Advances in Neural Information Processing Systems}.

\bibitem[{Bain et~al.(2021)Bain, Nagrani, Varol, and Zisserman}]{Bain21}
Max Bain, Arsha Nagrani, G{\"u}l Varol, and Andrew Zisserman. 2021.
\newblock Frozen in time: A joint video and image encoder for end-to-end retrieval.
\newblock In \emph{IEEE International Conference on Computer Vision}.

\bibitem[{Chen and Dolan(2011)}]{chen2011collecting}
David Chen and William~B Dolan. 2011.
\newblock Collecting highly parallel data for paraphrase evaluation.
\newblock In \emph{Annual meeting of the association for computational linguistics: human language technologies}.

\bibitem[{Chen et~al.(2023{\natexlab{a}})Chen, He, Guo, Zhu, Wang, Tang, and Liu}]{chen2023valor}
Sihan Chen, Xingjian He, Longteng Guo, Xinxin Zhu, Weining Wang, Jinhui Tang, and Jing Liu. 2023{\natexlab{a}}.
\newblock Valor: Vision-audio-language omni-perception pretraining model and dataset.
\newblock \emph{arXiv preprint arXiv:2304.08345}.

\bibitem[{Chen et~al.(2023{\natexlab{b}})Chen, Li, Wang, Zhao, Sun, Zhu, and Liu}]{chen2023vast}
Sihan Chen, Handong Li, Qunbo Wang, Zijia Zhao, Mingzhen Sun, Xinxin Zhu, and Jing Liu. 2023{\natexlab{b}}.
\newblock Vast: A vision-audio-subtitle-text omni-modality foundation model and dataset.
\newblock In \emph{Advances in Neural Information Processing Systems}.

\bibitem[{Chiang et~al.(2023)Chiang, Li, Lin, Sheng, Wu, Zhang, Zheng, Zhuang, Zhuang, Gonzalez, Stoica, and Xing}]{vicuna2023}
Wei-Lin Chiang, Zhuohan Li, Zi~Lin, Ying Sheng, Zhanghao Wu, Hao Zhang, Lianmin Zheng, Siyuan Zhuang, Yonghao Zhuang, Joseph~E. Gonzalez, Ion Stoica, and Eric~P. Xing. 2023.
\newblock Vicuna: An open-source chatbot impressing gpt-4 with 90\%* chatgpt quality.

\bibitem[{Chung et~al.(2022)Chung, Hou, Longpre, Zoph, Tay, Fedus, Li, Wang, Dehghani, Brahma et~al.}]{chung2022scaling}
Hyung~Won Chung, Le~Hou, Shayne Longpre, Barret Zoph, Yi~Tay, William Fedus, Eric Li, Xuezhi Wang, Mostafa Dehghani, Siddhartha Brahma, et~al. 2022.
\newblock Scaling instruction-finetuned language models.
\newblock \emph{arXiv preprint arXiv:2210.11416}.

\bibitem[{Dai et~al.(2023{\natexlab{a}})Dai, Li, Li, Tiong, Zhao, Wang, Li, Fung, and Hoi}]{dai2023instructblip}
Wenliang Dai, Junnan Li, Dongxu Li, Anthony Tiong, Junqi Zhao, Weisheng Wang, Boyang Li, Pascale Fung, and Steven Hoi. 2023{\natexlab{a}}.
\newblock Instruct{BLIP}: Towards general-purpose vision-language models with instruction tuning.
\newblock In \emph{Advances in Neural Information Processing Systems}.

\bibitem[{Dai et~al.(2023{\natexlab{b}})Dai, Li, Li, Tiong, Zhao, Wang, Li, Fung, and Hoi}]{instructblip}
Wenliang Dai, Junnan Li, Dongxu Li, Anthony Tiong, Junqi Zhao, Weisheng Wang, Boyang Li, Pascale Fung, and Steven Hoi. 2023{\natexlab{b}}.
\newblock Instruct{BLIP}: Towards general-purpose vision-language models with instruction tuning.
\newblock In \emph{Thirty-seventh Conference on Neural Information Processing Systems}.

\bibitem[{Diao et~al.(2023)Diao, Cheng, and Cheng}]{diao2023av}
Xingjian Diao, Ming Cheng, and Shitong Cheng. 2023.
\newblock Av-maskenhancer: Enhancing video representations through audio-visual masked autoencoder.
\newblock In \emph{International Conference on Tools with Artificial Intelligence}.

\bibitem[{Diao et~al.(2024)Diao, Zhang, Wu, Cheng, Ouyang, Wu, and Gui}]{diao2024learning}
Xingjian Diao, Chunhui Zhang, Tingxuan Wu, Ming Cheng, Zhongyu Ouyang, Weiyi Wu, and Jiang Gui. 2024.
\newblock Learning musical representations for music performance question answering.
\newblock In \emph{Findings of the Association for Computational Linguistics: EMNLP 2024}.

\bibitem[{Diao et~al.(2025)Diao, Zhang, Wu, Ouyang, Qing, Cheng, Vosoughi, and Gui}]{diao2025temporal}
Xingjian Diao, Chunhui Zhang, Weiyi Wu, Zhongyu Ouyang, Peijun Qing, Ming Cheng, Soroush Vosoughi, and Jiang Gui. 2025.
\newblock Temporal working memory: Query-guided segment refinement for enhanced multimodal understanding.
\newblock In \emph{Annual Conference of the North American Chapter of the Association for Computational Linguistics (NAACL 2025) Findings}.

\bibitem[{Fang et~al.(2023)Fang, Wang, Xie, Sun, Wu, Wang, Huang, Wang, and Cao}]{fang2022eva}
Yuxin Fang, Wen Wang, Binhui Xie, Quan Sun, Ledell Wu, Xinggang Wang, Tiejun Huang, Xinlong Wang, and Yue Cao. 2023.
\newblock Eva: Exploring the limits of masked visual representation learning at scale.
\newblock In \emph{IEEE/CVF Conference on Computer Vision and Pattern Recognition}.

\bibitem[{Fu et~al.(2023)Fu, Li, Gan, Lin, Wang, Wang, and Liu}]{fu2023empiricalmvm}
Tsu-Jui Fu, Linjie Li, Zhe Gan, Kevin Lin, William~Yang Wang, Lijuan Wang, and Zicheng Liu. 2023.
\newblock {An Empirical Study of End-to-End Video-Language Transformers with Masked Visual Modeling}.
\newblock In \emph{IEEE/CVF Conference on Computer Vision and Pattern Recognition}.

\bibitem[{Han et~al.(2024)Han, Jian, Hu, Liu, Wang, Fan, Ai, Huang, He, Yang, and You}]{han2024infimmwebmathb}
Xiaotian Han, Yiren Jian, Xuefeng Hu, Haogeng Liu, Yiqi Wang, Qihang Fan, Yuang Ai, Huaibo Huang, Ran He, Zhenheng Yang, and Quanzeng You. 2024.
\newblock Infi{MM}-webmath-40b: Advancing multimodal pre-training for enhanced mathematical reasoning.
\newblock In \emph{The 4th Workshop on Mathematical Reasoning and AI at NeurIPS'24}.

\bibitem[{He et~al.(2023)He, Chen, Ma, Huang, Jin, Liu, Fu, Yang, Liu, and Feng}]{he2023vlab}
Xingjian He, Sihan Chen, Fan Ma, Zhicheng Huang, Xiaojie Jin, Zikang Liu, Dongmei Fu, Yi~Yang, Jing Liu, and Jiashi Feng. 2023.
\newblock Vlab: Enhancing video language pre-training by feature adapting and blending.
\newblock \emph{arXiv preprint arXiv:2305.13167}.

\bibitem[{Jian et~al.(2023)Jian, Gao, and Vosoughi}]{jian2023bootstrapping}
Yiren Jian, Chongyang Gao, and Soroush Vosoughi. 2023.
\newblock Bootstrapping vision-language learning with decoupled language pre-training.
\newblock In \emph{Advances in Neural Information Processing Systems}.

\bibitem[{Jian et~al.(2024)Jian, Liu, Tao, Zhang, Vosoughi, and Yang}]{jian2024expedited}
Yiren Jian, Tingkai Liu, Yunzhe Tao, Chunhui Zhang, Soroush Vosoughi, and Hongxia Yang. 2024.
\newblock Expedited training of visual conditioned language generation via redundancy reduction.
\newblock In \emph{Proceedings of the 62nd Annual Meeting of the Association for Computational Linguistics (Volume 1: Long Papers)}.

\bibitem[{Kuo et~al.(2023)Kuo, Piergiovanni, Kim, xiyang luo, Caine, Li, Ogale, Zhou, Dai, Chen, Cui, and Angelova}]{kuo2023mammut}
Weicheng Kuo, AJ~Piergiovanni, Dahun Kim, xiyang luo, Benjamin Caine, Wei Li, Abhijit Ogale, Luowei Zhou, Andrew~M. Dai, Zhifeng Chen, Claire Cui, and Anelia Angelova. 2023.
\newblock Ma{MMUT}: A simple architecture for joint learning for multimodal tasks.
\newblock \emph{Transactions on Machine Learning Research}.

\bibitem[{Li et~al.(2023{\natexlab{a}})Li, Li, Le, Wang, Savarese, and Hoi}]{li2022lavis}
Dongxu Li, Junnan Li, Hung Le, Guangsen Wang, Silvio Savarese, and Steven~C.H. Hoi. 2023{\natexlab{a}}.
\newblock {LAVIS}: A one-stop library for language-vision intelligence.
\newblock In \emph{Annual Meeting of the Association for Computational Linguistics: System Demonstrations}.

\bibitem[{Li et~al.(2022{\natexlab{a}})Li, Li, Li, Niebles, and Hoi}]{li2021alignprompt}
Dongxu Li, Junnan Li, Hongdong Li, Juan~Carlos Niebles, and Steven~C.H. Hoi. 2022{\natexlab{a}}.
\newblock Align and prompt: Video-and-language pre-training with entity prompts.
\newblock In \emph{IEEE/CVF Conference on Computer Vision and Pattern Recognition}.

\bibitem[{Li et~al.(2023{\natexlab{b}})Li, Li, Savarese, and Hoi}]{li2023blip}
Junnan Li, Dongxu Li, Silvio Savarese, and Steven Hoi. 2023{\natexlab{b}}.
\newblock Blip-2: Bootstrapping language-image pre-training with frozen image encoders and large language models.
\newblock In \emph{International conference on machine learning}.

\bibitem[{Li et~al.(2022{\natexlab{b}})Li, Li, Xiong, and Hoi}]{li2022blip}
Junnan Li, Dongxu Li, Caiming Xiong, and Steven Hoi. 2022{\natexlab{b}}.
\newblock Blip: Bootstrapping language-image pre-training for unified vision-language understanding and generation.
\newblock In \emph{International Conference on Machine Learning}.

\bibitem[{Li et~al.(2023{\natexlab{c}})Li, He, Wang, Li, Wang, Luo, Wang, Wang, and Qiao}]{2023videochat}
Kunchang Li, Yinan He, Yi~Wang, Yizhuo Li, Wenhai Wang, Ping Luo, Yali Wang, Limin Wang, and Yu~Qiao. 2023{\natexlab{c}}.
\newblock Videochat: Chat-centric video understanding.
\newblock \emph{arXiv preprint arXiv:2305.06355}.

\bibitem[{Liang et~al.(2023)Liang, Zhu, Wang, and Yang}]{liang2023icocap}
Yuanzhi Liang, Linchao Zhu, Xiaohan Wang, and Yi~Yang. 2023.
\newblock Icocap: Improving video captioning by compounding images.
\newblock \emph{IEEE Transactions on Multimedia}.

\bibitem[{Lin et~al.(2014)Lin, Maire, Belongie, Hays, Perona, Ramanan, Doll{\'a}r, and Zitnick}]{lin2014microsoft}
Tsung-Yi Lin, Michael Maire, Serge Belongie, James Hays, Pietro Perona, Deva Ramanan, Piotr Doll{\'a}r, and C~Lawrence Zitnick. 2014.
\newblock Microsoft coco: Common objects in context.
\newblock In \emph{European Conference on Computer Vision}.

\bibitem[{Liu et~al.(2024)Liu, You, Wang, Han, Zhai, Liu, Chen, Jian, Tao, Yuan, He, and Yang}]{liu-etal-2024-infimm}
Haogeng Liu, Quanzeng You, Yiqi Wang, Xiaotian Han, Bohan Zhai, Yongfei Liu, Wentao Chen, Yiren Jian, Yunzhe Tao, Jianbo Yuan, Ran He, and Hongxia Yang. 2024.
\newblock {I}nfi{MM}: Advancing multimodal understanding with an open-sourced visual language model.
\newblock In \emph{Findings of the Association for Computational Linguistics: ACL 2024}.

\bibitem[{Liu et~al.(2023{\natexlab{a}})Liu, Li, Li, and Lee}]{liu2023improvedllava}
Haotian Liu, Chunyuan Li, Yuheng Li, and Yong~Jae Lee. 2023{\natexlab{a}}.
\newblock Improved baselines with visual instruction tuning.

\bibitem[{Liu et~al.(2023{\natexlab{b}})Liu, Li, Wu, and Lee}]{liu2023llava}
Haotian Liu, Chunyuan Li, Qingyang Wu, and Yong~Jae Lee. 2023{\natexlab{b}}.
\newblock Visual instruction tuning.
\newblock In \emph{NeurIPS}.

\bibitem[{Luo et~al.(2023)Luo, Zhao, Yang, Dong, Qiu, Lu, Wang, and Wei}]{luo2023valley}
Ruipu Luo, Ziwang Zhao, Min Yang, Junwei Dong, Minghui Qiu, Pengcheng Lu, Tao Wang, and Zhongyu Wei. 2023.
\newblock Valley: Video assistant with large language model enhanced ability.
\newblock \emph{arXiv preprint arXiv:2306.07207}.

\bibitem[{Muhammad~Maaz and Khan(2023)}]{Maaz2023VideoChatGPT}
Salman~Khan Muhammad~Maaz, Hanoona~Rasheed and Fahad Khan. 2023.
\newblock Video-chatgpt: Towards detailed video understanding via large vision and language models.
\newblock \emph{ArXiv 2306.05424}.

\bibitem[{Ordonez et~al.(2011)Ordonez, Kulkarni, and Berg}]{ordonez2011im2text}
Vicente Ordonez, Girish Kulkarni, and Tamara Berg. 2011.
\newblock Im2text: Describing images using 1 million captioned photographs.
\newblock In \emph{Advances in neural information processing systems}.

\bibitem[{Rennie et~al.(2017)Rennie, Marcheret, Mroueh, Ross, and Goel}]{rennie2017self}
Steven~J Rennie, Etienne Marcheret, Youssef Mroueh, Jerret Ross, and Vaibhava Goel. 2017.
\newblock Self-critical sequence training for image captioning.
\newblock In \emph{IEEE conference on computer vision and pattern recognition}.

\bibitem[{Sarto et~al.(2023)Sarto, Barraco, Cornia, Baraldi, and Cucchiara}]{sarto2023positive}
Sara Sarto, Manuele Barraco, Marcella Cornia, Lorenzo Baraldi, and Rita Cucchiara. 2023.
\newblock Positive-augmented contrastive learning for image and video captioning evaluation.
\newblock In \emph{Proceedings of the IEEE/CVF conference on computer vision and pattern recognition}.

\bibitem[{Schuhmann et~al.(2022)Schuhmann, Beaumont, Vencu, Gordon, Wightman, Cherti, Coombes, Katta, Mullis, Wortsman et~al.}]{schuhmannlaion}
Christoph Schuhmann, Romain Beaumont, Richard Vencu, Cade~W Gordon, Ross Wightman, Mehdi Cherti, Theo Coombes, Aarush Katta, Clayton Mullis, Mitchell Wortsman, et~al. 2022.
\newblock Laion-5b: An open large-scale dataset for training next generation image-text models.
\newblock In \emph{Advances in Neural Information Processing Systems: Datasets and Benchmarks Track}.

\bibitem[{Sharma et~al.(2018)Sharma, Ding, Goodman, and Soricut}]{sharma2018conceptual}
Piyush Sharma, Nan Ding, Sebastian Goodman, and Radu Soricut. 2018.
\newblock Conceptual captions: A cleaned, hypernymed, image alt-text dataset for automatic image captioning.
\newblock In \emph{Annual Meeting of the Association for Computational Linguistics}.

\bibitem[{Su et~al.(2023)Su, Lan, Li, Xu, Wang, and Cai}]{su2023pandagpt}
Yixuan Su, Tian Lan, Huayang Li, Jialu Xu, Yan Wang, and Deng Cai. 2023.
\newblock Pandagpt: One model to instruction-follow them all.
\newblock In \emph{Workshop on Taming Large Language Models: Controllability in the era of Interactive Assistants}.

\bibitem[{Wang et~al.(2022{\natexlab{a}})Wang, Yang, Hu, Li, Lin, Gan, Liu, Liu, and Wang}]{wang2022git}
Jianfeng Wang, Zhengyuan Yang, Xiaowei Hu, Linjie Li, Kevin Lin, Zhe Gan, Zicheng Liu, Ce~Liu, and Lijuan Wang. 2022{\natexlab{a}}.
\newblock {GIT}: A generative image-to-text transformer for vision and language.
\newblock \emph{Transactions on Machine Learning Research}.

\bibitem[{Wang et~al.(2019)Wang, Wu, Chen, Li, Wang, and Wang}]{wang2019vatex}
Xin Wang, Jiawei Wu, Junkun Chen, Lei Li, Yuan-Fang Wang, and William~Yang Wang. 2019.
\newblock Vatex: A large-scale, high-quality multilingual dataset for video-and-language research.
\newblock In \emph{IEEE/CVF international conference on computer vision}.

\bibitem[{Wang et~al.(2022{\natexlab{b}})Wang, Yu, Yu, Dai, Tsvetkov, and Cao}]{wang2022simvlm}
Zirui Wang, Jiahui Yu, Adams~Wei Yu, Zihang Dai, Yulia Tsvetkov, and Yuan Cao. 2022{\natexlab{b}}.
\newblock Sim{VLM}: Simple visual language model pretraining with weak supervision.
\newblock In \emph{International Conference on Learning Representations}.

\bibitem[{Xu et~al.(2023)Xu, Ye, Yan, Shi, Ye, Xu, Li, Bi, Qian, Wang, Xu, Zhang, Huang, Huang, and Zhou}]{mplug2}
Haiyang Xu, Qinghao Ye, Ming Yan, Yaya Shi, Jiabo Ye, Yuanhong Xu, Chenliang Li, Bin Bi, Qi~Qian, Wei Wang, Guohai Xu, Ji~Zhang, Songfang Huang, Fei Huang, and Jingren Zhou. 2023.
\newblock mplug-2: a modularized multi-modal foundation model across text, image and video.
\newblock In \emph{International Conference on Machine Learning}.

\bibitem[{Xu et~al.(2021)Xu, Ghosh, Huang, Okhonko, Aghajanyan, and Feichtenhofer}]{xu2021videoclip}
Hu~Xu, Gargi Ghosh, Po-Yao~(Bernie) Huang, Dmytro Okhonko, Armen Aghajanyan, and Florian Metze Luke Zettlemoyer~Christoph Feichtenhofer. 2021.
\newblock Videoclip: Contrastive pre-training for zero-shot video-text understanding.
\newblock In \emph{Conference on Empirical Methods in Natural Language Processing}.

\bibitem[{Xu et~al.(2016)Xu, Mei, Yao, and Rui}]{xu2016msr}
Jun Xu, Tao Mei, Ting Yao, and Yong Rui. 2016.
\newblock Msr-vtt: A large video description dataset for bridging video and language.
\newblock In \emph{IEEE conference on computer vision and pattern recognition}.

\bibitem[{Yan et~al.(2022)Yan, Zhu, Wang, Cao, Zhang, Ghosh, Wu, and Yu}]{yan2022videotext}
Shen Yan, Tao Zhu, Zirui Wang, Yuan Cao, Mi~Zhang, Soham Ghosh, Yonghui Wu, and Jiahui Yu. 2022.
\newblock \href {https://arxiv.org/abs/2212.04979} {Videococa: Video-text modeling with zero-shot transfer from contrastive captioners}.
\newblock \emph{Preprint}, arXiv:2212.04979.

\bibitem[{Yang et~al.(2021)Yang, Bisk, and Gao}]{yang2021}
J.~Yang, Y.~Bisk, and J.~Gao. 2021.
\newblock Taco: Token-aware cascade contrastive learning for video-text alignment.
\newblock In \emph{International Conference on Computer Vision}.

\bibitem[{Yang et~al.(2024)Yang, Dai, Liu, Wang, Jiang, Tian, and Zhang}]{yang2024cliperase}
Tianyu Yang, Lisen Dai, Zheyuan Liu, Xiangqi Wang, Meng Jiang, Yapeng Tian, and Xiangliang Zhang. 2024.
\newblock Cliperase: Efficient unlearning of visual-textual associations in clip.
\newblock \emph{arXiv preprint arXiv:2410.23330}.

\bibitem[{Yu et~al.(2022)Yu, Wang, Vasudevan, Yeung, Seyedhosseini, and Wu}]{yu2022coca}
Jiahui Yu, Zirui Wang, Vijay Vasudevan, Legg Yeung, Mojtaba Seyedhosseini, and Yonghui Wu. 2022.
\newblock Coca: Contrastive captioners are image-text foundation models.
\newblock \emph{Transactions on Machine Learning Research}.

\bibitem[{Zellers et~al.(2021)Zellers, Lu, Hessel, Yu, Park, Cao, Farhadi, and Choi}]{zellers2021merlot}
Rowan Zellers, Ximing Lu, Jack Hessel, Youngjae Yu, Jae~Sung Park, Jize Cao, Ali Farhadi, and Yejin Choi. 2021.
\newblock {MERLOT}: Multimodal neural script knowledge models.
\newblock In \emph{Advances in Neural Information Processing Systems}.

\bibitem[{Zhang et~al.(2022{\natexlab{a}})Zhang, Huang, Li, Zhang, Ye, and Zhang}]{zhang2022look}
Chunhui Zhang, Chao Huang, Youhuan Li, Xiangliang Zhang, Yanfang Ye, and Chuxu Zhang. 2022{\natexlab{a}}.
\newblock Look twice as much as you say: Scene graph contrastive learning for self-supervised image caption generation.
\newblock In \emph{Proceedings of the 31st ACM International Conference on Information \& Knowledge Management}.

\bibitem[{Zhang et~al.(2024)Zhang, Jian, Ouyang, and Vosoughi}]{zhang-etal-2024-working}
Chunhui Zhang, Yiren Jian, Zhongyu Ouyang, and Soroush Vosoughi. 2024.
\newblock Working memory identifies reasoning limits in language models.
\newblock In \emph{Proceedings of the 2024 Conference on Empirical Methods in Natural Language Processing}.

\bibitem[{Zhang et~al.(2023)Zhang, Li, and Bing}]{damonlpsg2023videollama}
Hang Zhang, Xin Li, and Lidong Bing. 2023.
\newblock Video-{LL}a{MA}: An instruction-tuned audio-visual language model for video understanding.
\newblock In \emph{Empirical Methods in Natural Language Processing: System Demonstrations}.

\bibitem[{Zhang et~al.(2022{\natexlab{b}})Zhang, Roller, Goyal, Artetxe, Chen, Chen, Dewan, Diab, Li, Lin et~al.}]{zhang2022opt}
Susan Zhang, Stephen Roller, Naman Goyal, Mikel Artetxe, Moya Chen, Shuohui Chen, Christopher Dewan, Mona Diab, Xian Li, Xi~Victoria Lin, et~al. 2022{\natexlab{b}}.
\newblock Opt: Open pre-trained transformer language models.
\newblock \emph{arXiv preprint arXiv:2205.01068}.

\bibitem[{Zhu et~al.(2024)Zhu, Chen, Shen, Li, and Elhoseiny}]{zhu2023minigpt}
Deyao Zhu, Jun Chen, Xiaoqian Shen, Xiang Li, and Mohamed Elhoseiny. 2024.
\newblock Mini{GPT}-4: Enhancing vision-language understanding with advanced large language models.
\newblock In \emph{International Conference on Learning Representations}.

\end{thebibliography}

\appendix
\section{Related Work and Background}
\label{app:sec:related-work}
\paragraph{Image-Text Models}
Large-scale pretraining has revolutionized the field of image-text models, enabling significant advances. Models such as CoCa~\citep{yu2022coca} and SimVLM~\cite{wang2022simvlm}, which are trained from scratch on billions of image-text pairs, have set new benchmarks in generative tasks such as open-ended visual question answering (VQA) and visual captioning. BLIP-2 addresses the computational demands of pretraining from scratch by reusing existing pre-trained parameters from Vision Transformer (ViT) and LLMs and integrating them with a frozen pre-trained state. A key innovation in BLIP-2 is the introduction of the Q-former connector, carefully designed to enhance the interaction between visual and language modalities \citep{li2023blip}. This methodology has inspired subsequent innovations in visual-lingual tuning, with newer models often incorporating the pre-trained Q-former alongside the \texttt{eva-vit-g} model from BLIP-2, demonstrating the lasting impact of this methodology \citep{instructblip, zhu2023minigpt, yang2024cliperase, 2023videochat}.

\paragraph{Video-Text Models}
Video-text models typically extend the capabilities of image-text models by integrating temporal feature aggregation to capture dynamic content, as exemplified by VideoCoCa \citep{yan2022videotext}. In addition, specialized models such as Video-LLaMA enhance the processing of temporal dynamics by embedding multiple temporal Q-former layers, facilitating nuanced interactions across modalities. Such advances refine the synergy between video Q-formers and LLMs within the model architecture, building on the foundation of BLIP-2 \citep{damonlpsg2023videollama}. Building on these developments, recent studies, including VideoChat, PandaGPT, Valley, and Video-ChatGPT, investigate the embedding of frozen LLMs into video LMs, pushing the boundaries of the field \citep{2023videochat, su2023pandagpt, luo2023valley, Maaz2023VideoChatGPT}. In our study, we use BLIP-2 as a basic model for captioning, first pre-trained on images and then adapted to video by incorporating a video frame merging mechanism that effectively captures temporal nuances. This simplicity allows us to focus on evaluating the effects of model size, data volume, and training strategies on video captioning performance as we scale. 

\paragraph{Difference between Image and Video Captioning}
The fundamental difference between image and video annotation stems from their source inputs: image annotation processes a single static image, while video annotation requires an understanding of the temporal dynamics over a sequence of frames. When adapted to video, pre-trained image models such as GIT~\citep{wang2022git}, VideoCoCa~\citep{yan2022videotext}, and IcoCap~\citep{liang2023icocap} show remarkable adaptability to video with only moderate modifications, demonstrating their transferability. Conversely, video-specific models, including Video-LLaMA~\citep{damonlpsg2023videollama} and VideoChat \citep{2023videochat}, use different sampling techniques to effectively capture temporal dynamics. Furthermore, models such as ALPRO \citep{li2021alignprompt} and VIOLET \citep{fu2023empiricalmvm} utilize extensive web-crawled datasets to achieve end-to-end training, enriching their learning process. In our study, instead of emulating the complex adaptations typical of specialized video models, we adopt a streamlined approach that uses averaging or concatenation to merge temporal information from sampled video frames. This method allows us to focus on evaluating the effects of model size, data volume, and training strategies on video captioning performance as we scale.

\section{Preliminary}
\label{app:sec:preliminary}
To effectively analyze the impact of specialized video adaptations without the confounding effects of architectural design variations, we base our methodology on BLIP-2, a basic image captioning model. We then describe the rationale for selecting BLIP-2 for our study.

\paragraph{Architecture of BLIP-2}
BLIP-2 is originally designed to convert images into captions through a simple pipeline consisting of three main components: Vision, Connector, and Language:
\underline{\textbf{\textit{(i)} Vision}} ViT serves as the entry into the BLIP-2 architecture, encoding images into a series of visual tokens. For example, a 224$\times$224 image is transformed into 256 different visual tokens, laying the foundation for subsequent processing;
\underline{\textbf{\textit{(ii)} Modal connector}} Q-former, positioned between ViT and LLM, bridges the gap between visual and language modalities. Its primary function is to project the sequence of the visual tokens generated by the ViT into a format compatible with language processing. A distinctive feature of the Q-former is its ability to condense the visual token array to a predetermined size, typically 32 tokens, regardless of the original number. This token reduction is not simply a numerical compression, but involves a sophisticated transformation into a language modality, resulting in so-called \textit{soft prompts}. These soft prompts, now in tensor form, are then passed to the LLM for caption generation;
\underline{\textbf{\textit{(iii)} Language}} LLM is responsible for generating the textual captions. It interprets the soft prompts from the Q-former and weaves them into a coherent caption that accurately reflects the visual content. This step is the culmination of the BLIP-2 pipeline, which transforms visual input into descriptive language.

\paragraph{Rationale for Choosing BLIP-2 as the Base Model}
In the field of vision language generative learning, many pre-trained image-based vision LMs are possible candidates besides BLIP-2, such as the LLaVA series, miniGPT-4, OpenCoCa, and OpenFlamingo, each offering different capabilities and features. Given the wide range of options available, our selection of pre-trained BLIP-2 is guided by specific criteria:

\underline{\textit{First}}, LLaVA uses a linear projection layer to project visual tokens from ViT and then feeds the projected tokens into LLMs. However, this linear projection layer keeps the visual tokens \textit{consistent}, which means that this connector does not compress the visual token into fewer numbers. Although this redundant representation format does not meet the efficiency bottleneck on a single image as we extend the input modality to a single video containing multiple frames, it may exhaust the maximum token length capacity of an LLM. In contrast, BLIP-2 can reduce the number of tokens for each image/frame to a fixed number (e.g., 32). This efficient design avoids placing additional significant demands on the token length capacity of an LLM. 
\underline{\textit{Second}}, mini-GPT4, an instruction-tuned BLIP-2, also uses a linear projection layer to project visual tokens from ViT and then feeds the projected tokens into LLMs. Therefore, it also faces a similar limitation as LLaVA: when processing video frames, mini-GPT4's LLM token capacity also quickly hits a forward-backward bottleneck, limiting the number of frames that can be effectively captioned. 
\underline{\textit{Third}}, while Flamingo is easily adapted to video data due to its cross-modal attention design, its open-source reproduction, OpenFlamingo, underperforms BLIP-2 according to \citet{li2023blip}'s experiments.
Third, Flamingo's design, which features cross-modal attention, facilitates its straightforward adaptation to video data; however, experiments conducted by \citet{li2023blip} imply that OpenFlamingo, an open-source version of Flamingo, does not perform as well as BLIP-2.
Therefore, compared to LLaVA and mini-GPT4, BLIP-2 can be easily applied to video data to process multiple frames by averaging or concatenating the tokens of multiple frames (with a short length for the token of each frame, e.g. 32 tokens). We find that the BLIP-2 is characterized by its generality and simplicity, making it particularly well suited to the task of video captioning. Its design allows for minimal modification, allowing us to focus on the core factors that contribute to the effectiveness of video captioning models. This strategic choice is consistent with our goal of isolating and understanding the key elements that drive effective video captioning.

\section{Additional Experimental Details}
\label{app:exp}
\subsection{ Setup}
\paragraph{Video Dataset Overview}
Our study uses the \textbf{MSR-VTT} dataset~\citep{xu2016msr}, a comprehensive open-domain video captioning resource. It includes 10,000 video clips across 20 different categories, with each clip annotated with 20 unique English sentences by contributors via Amazon Mechanical Turk. The dataset contains approximately 29,000 different words within the captions. For our experiments, we adhere to the conventional dataset partitioning: 6,513 clips for training, 497 for validation, and 2,990 for testing.

\paragraph{Training Configuration}
Training is conducted on eight NVIDIA RTX A6000 GPUs, utilizing the MSR-VTT dataset. Optimization is performed using the AdamW algorithm, with a setup that includes a weight decline of 0.05, an initial learning rate of $5 \times 10^{-5}$, and a minimum learning rate of $1 \times 10^{-5}$. The models are trained with a batch size of 32 over 32 epochs, with learning rate adjustments governed by a cosine annealing scheduler. 

\subsection{Model Information}
\label{sec:model-list}
Our video captioning model uses the image pre-trained BLIP-2 as its foundation. The BLIP-2 model itself is initially trained from scratch using the MSCOCO~\citep{lin2014microsoft} and CapFilt~\citep{li2022blip} datasets, with additional data from the pseudo-labeled Conceptual Captioning~\citep{sharma2018conceptual}, SBU~\citep{ordonez2011im2text}, and LAION~\citep{schuhmannlaion} collections. 
\noindent Our study employs ViT (\texttt{eva-vit-g} released from \citep{fang2022eva}) due to its proven effectiveness. In the realm of LM decoders, we investigate the capabilities of \texttt{OPT}~\citep{zhang2022opt}, \texttt{Flan-T5}~\citep{chung2022scaling}, and \texttt{vicuna-7b}~\citep{vicuna2023}, as the large pretrained LM decoders have shown their capabilities~\cite{zhang-etal-2024-working}. To adapt BLIP-2 for video, we utilize \texttt{bert-base-uncased} for the q-former architecture, maintaining parameter consistency with the image-trained version of BLIP-2. Additionally, we implement a frame token concatenation mechanism for aggregating temporal information from videos without increasing the parameter count.
We provide the detailed structures, pre-train data, and language backbones in Tab.~\ref{tab:model-list}.
\begin{table*}[h]
\centering
\resizebox{0.8\textwidth}{!}{
\begin{tblr}{
  hline{1,11} = {-}{0.08em},
  hline{2} = {-}{0.05em},
}
Model    & {\# pretrain \\image-text} & {\#video-text} & {Vision \\Backbone} & {Language \\Backbone} \\
IcoCap~\citep{liang2023icocap}    & -                          & -                          & CLIP-V              & Transformer           \\
MaMMUT~\citep{kuo2023mammut}    & 1.8B                       & -                          & ViT                 & Transformer           \\
VideoCoCa~\citep{yan2022videotext} & 3B                          & 136M+8.7M                  & CoCa-V              & CoCa-T                \\
VALOR~\citep{chen2023valor}     & 1.18M                      & 1.18M                      & CLIP-V/VideoSwin    & BERT                  \\
VLAB~\citep{he2023vlab}      & 5M+12M                     & 10.7M                      & ViT giant           & Transformer           \\
GIT2~\citep{wang2022git}      & 12.9B                      & -                          & CoSwin              & Transformer           \\
VAST~\cite{chen2023vast}      & -                          & 27M                        & ViT                 & BERT                  \\
mPLUG-2~\citep{mplug2}   & 14M                        & 2.5M                       & ViT-L/14            & BERT-L                \\
Ours      & 129M                          & 6K                          & EVA-ViT-G                   & Flan-T5-XL                     
\end{tblr}}
\caption{The number of pre-train image-text and video-text pairs, vision backbone, and the language backbone for each video captioning model. }
\label{tab:model-list}
\end{table*}

\section{Training Analysis and Results on Other Datasets}
\label{app:ana}
\subsection{Model Scale}
\label{sec:app:1}
\begin{figure*}[!t]
\centering
\includegraphics[width=1\textwidth]{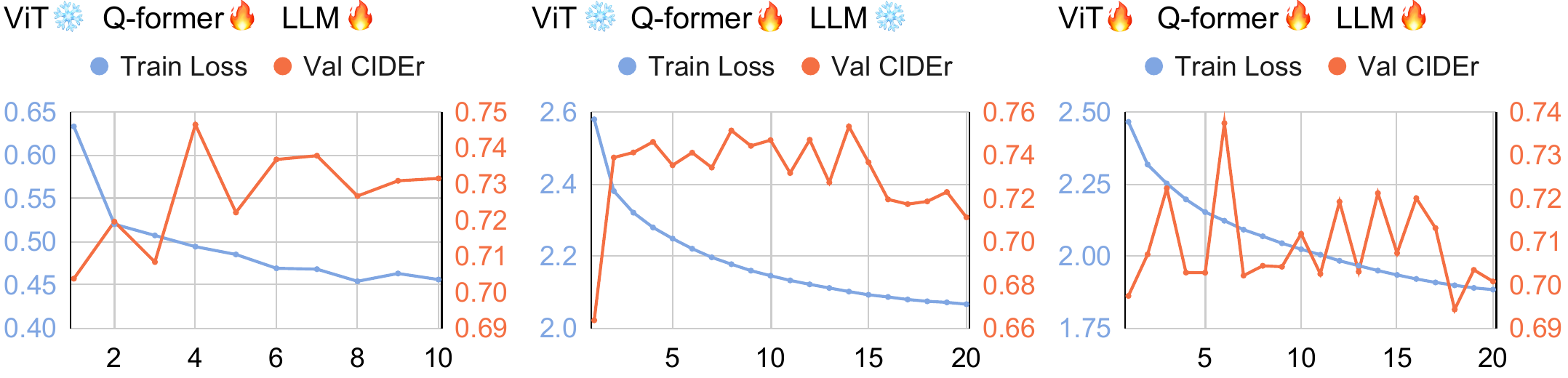}
\caption{Training curves of the video captioning model on MSR-VTT, with different module freezing configurations. The vision backbone is ViT, and the language backbone is FLAN-T5. The curves represent three settings: (a) ViT frozen, (b) only Q-former trainable, and (c) all components trainable.}
\label{fig:app-1}
\end{figure*}
\subsubsection{Trainability: modal connector $>$ LLM $>$ ViT}
\label{app:sec:qformer-llm-vit}
Fig.~\ref{fig:app-1} presents the training curves of the video captioning model on MSR-VTT for different module freezing configurations: (a) ViT frozen, (b) only Q-Former trainable, and (c) all components trainable. The curves highlight the differences in trainability between the modal connector (Q-Former), the LLM, and the vision transformer (ViT).

The training curves indicate that setting (b), \textbf{where only the Q-Former is trainable, shows the most stable performance, reaching peak validation CIDEr at epoch 14 without significant overfitting.} In contrast, when additional components are trainable—such as the LLM in setting (c) or the ViT in setting (a)—the models reach peak performance earlier, at 6 and 4 epochs, respectively, but exhibit rapid overfitting afterward. This pattern suggests that increasing the number of trainable components complicates the optimization process, leading to quicker convergence but also accelerated overfitting. Consequently, setting (b) achieves the highest test CIDEr score (73.6), followed by setting (c) (73.0), and setting (a) (68.4).

\textbf{Training the LLM also proves to be effective for video captioning}, as reflected by the higher CIDEr score in setting (c). LLMs benefit from extensive pre-training on structured text, which enhances their ability to reason and assemble concepts. This capability allows them to align seamlessly with other modalities and reorganize visual inputs into coherent captions, making them a crucial component for video captioning tasks.

\textbf{In contrast, training the ViT module appears suboptimal (or even counterproductive) for video captioning}, as shown by the lower performance in setting (a). While large-scale pre-trained vision models like CLIP can capture fine-grained visual details, they often lack the structured representations necessary for composing visual information into coherent descriptions. This limitation affects the ability of the model to generate accurate captions when the ViT is a primary trainable component.

\subsubsection{Mid-sized LLMs offer trainability for video captioning}
\label{app:sec:mid-size-llm}
\begin{figure*}[!t]
\centering
\includegraphics[width=1\textwidth]{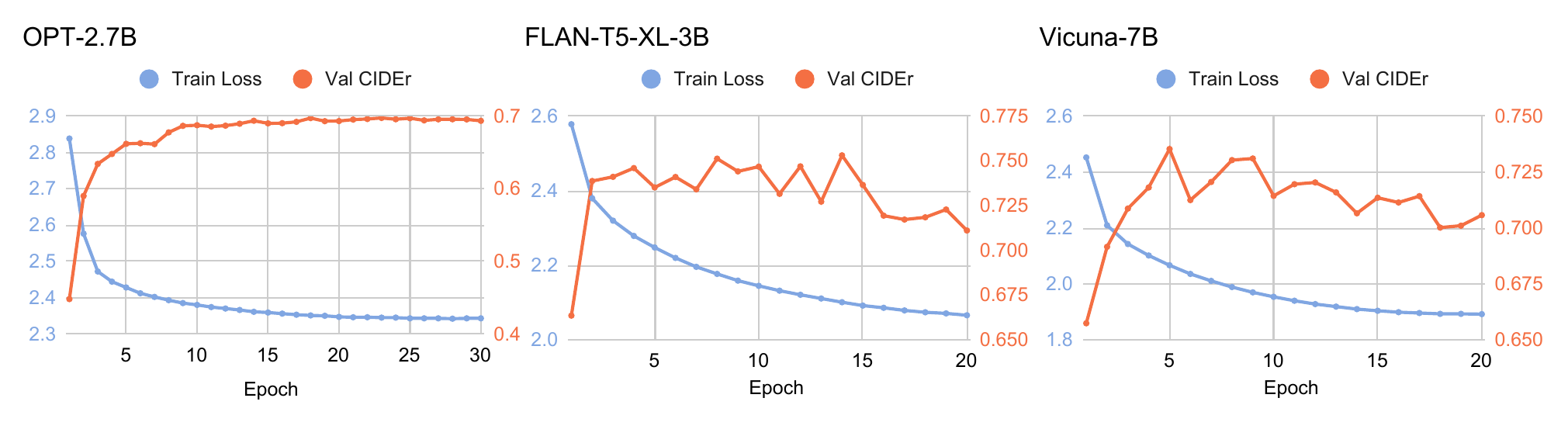}
\caption{Training curves of a video captioning model with different sizes of LLMs. (a), (b), and (c) show training curves of LLMs with sizes 2.7B, 3B, and 7B respectively.}
\label{fig:app:small2large}
\end{figure*}
To validate the advantages of mid-sized LLMs, we present the training dynamics for three different LM sizes in Fig.~\ref{fig:app:small2large}. The training curves indicate that larger models converge more quickly: OPT-2.7B requires 20 epochs to reach peak performance, Flan-T5-XL-3B takes 14 epochs, and Vicuna-7B converges in just 5 epochs. Although OPT-2.7B undergoes the longest training process, it fails to overfit the data, indicating limited model complexity. In contrast, both Flan-T5-XL-3B and Vicuna-7B show signs of overfitting soon after reaching peak performance, reflecting their greater model expressiveness for the video captioning task.

\textbf{Flan-T5-XL-3B, with fewer parameters than Vicuna-7B, demonstrates sufficient complexity for video captioning tasks while requiring less computational power.} Its moderate size avoids the additional burden of excessive parameters, leading to a more balanced and efficient learning process. In conclusion, mid-sized LMs, such as \textbf{Flan-T5-XL-3B, provide the optimal balance of trainability and complexity for video captioning, offering more efficient learning and better performance compared to their larger counterparts.}

\subsection{Data Efficiency}
\label{sec:app:2}
\begin{figure}[!t]
\centering
\includegraphics[width=0.5\textwidth]{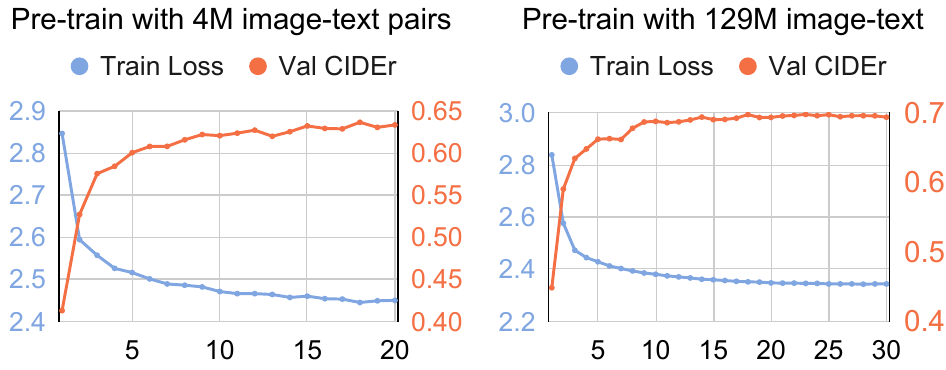}
\caption{Training curve of a video captioning model with different sizes of pre-trained image-text pairs. (a) and (b) show training curves of models pre-trained with 4M and 129M image-text pairs respectively.}
\label{fig:app-3}
\end{figure}
\subsubsection{Image-Text pretraining offers transferability to video tasks}
\label{app:sec:image-text-pretrain}
Fig.~\ref{fig:app-3} illustrates that BLIP-2, \textbf{when pre-trained on a larger image-text dataset (129M pairs, officially released by the BLIP-2 group), converges faster and achieves a higher performance limit compared to the model trained with 4M image-text pairs.} This difference suggests that video captioning, while not as demanding in reasoning as tasks like VQA, still requires a strong ability to understand and describe visual content accurately. Extensive exposure to large-scale image-text data significantly improves the model’s grounding process, enabling it to better understand and articulate visual content in video tasks. Thus, pre-training on extensive image-text datasets enhances the model’s ability to map visual concepts from the vision domain to the language domain, making it more effective for video captioning. These results further highlight the \textit{effectiveness} of reusing extensively pre-trained image-text models for video captioning tasks.

\subsubsection{Lower resolution efficiently supports video captioning}
\label{app:sec:lower-resolution}
\begin{figure}[!t]
\centering
\includegraphics[width=0.5\textwidth]{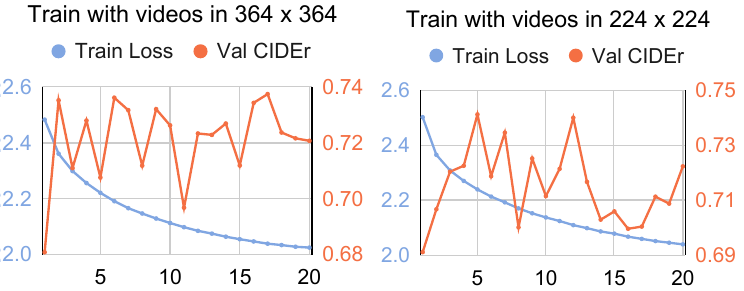}
\caption{The training dynamics of a video captioning model with videos in different resolutions. (a) and (b) shows training curves of models trained with videos in 364$\times$364 (up-sampling from original resolution 320x240 from MSR-VTT) and 224$\times$224 respectively.}
\label{fig:app-6}
\end{figure}
Fig.~\ref{fig:app-6} compares the training dynamics of models using different video resolutions, showing that higher resolution videos (364$\times$364) exhibit slightly more stable performance when combined with a stronger frame aggregator. \textbf{However, when the video frame aggregator is not highly sophisticated, lower resolution (224$\times$224) proves to be efficient and effective, providing sufficient visual information for the model to perceive and generate accurate captions.} These findings indicate that lower resolution is not only sufficient but also more efficient for video captioning, especially when using basic frame aggregation techniques.

\begin{figure}[!t]
\centering
\includegraphics[width=0.5\textwidth]{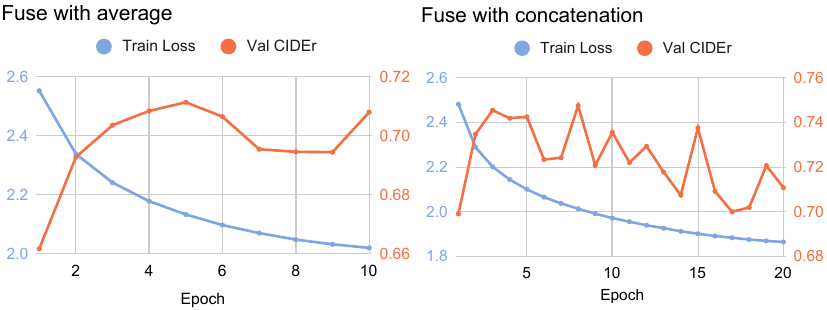}
\caption{The training dynamics of a video captioning model with different fusion mechanisms for video frames. (a) and (b) show training curves of models that adopt the average and concatenation mechanisms respectively.}
\label{fig:app-5}
\end{figure}
\subsubsection{Frame concatenation effectively captures temporality}
\label{app-5-frame}
Fig.~\ref{fig:app-5} illustrates the training dynamics for two fusion mechanisms: frame concatenation and averaging. \textbf{The model using concatenation reaches peak validation performance at epoch 8, suggesting that the complex visual tokens retain sufficient temporal information for effective learning.} In contrast, the averaging mechanism demonstrates weaker performance, with significant oscillations after epoch 5, indicating that it fails to provide enough temporal information for stable training. These results indicate that \textbf{frame concatenation is essential for effectively preserving temporal information, making it a more suitable approach for capturing visual concepts in video captioning.}

\subsection{Training Supervision}
\label{sec:app:3}
\begin{figure*}[!t]
\centering
\includegraphics[width=1\textwidth]{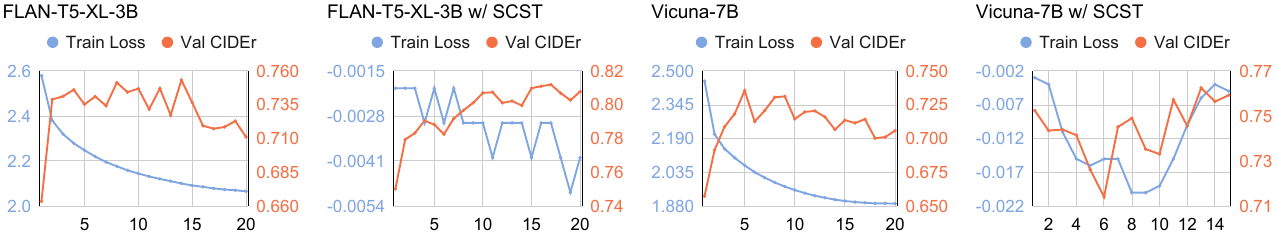}
\caption{The training dynamics for the model when trained with/without SCST in LLM.}
\label{fig:app-4}
\end{figure*}
\subsubsection{Reinforcement learning aligns captioning with human preference}
Fig.~\ref{fig:app-4} shows the training dynamics for the Flan-T5-XL-3B and Vicuna-7B models with and without Self-Critical Sequence Training (SCST). The plots illustrate how SCST affects the relationship between training loss and validation CIDEr score. When SCST is applied, the training loss shows more variation, but the validation CIDEr score remains higher compared to models without SCST. For example, Flan-T5-XL-3B with SCST achieves a validation CIDEr score of about 0.82 despite increasing training loss, while Vicuna-7B with SCST maintains a CIDEr score of about 0.77.

Without SCST, both models follow a more conventional pattern where a steady decrease in training loss corresponds to a plateau in validation performance. In contrast, SCST introduces a decoupling effect: \textbf{fluctuations in training loss are no longer directly correlated with changes in validation CIDEr, suggesting that SCST promotes learning focused on optimizing human-centered metrics.} These results show that reinforcement learning via SCST effectively aligns the training process with human evaluation standards, prioritizing high-quality label generation that aligns with human judgment over simply minimizing training loss.

\subsection{Experiments on MSVD and VATEX dataset}
\label{sec:app:msvd}
\begin{figure*}[t]
    \centering
    \begin{subfigure}[t]{0.2455\textwidth}
        \centering
        \includegraphics[width=\textwidth]{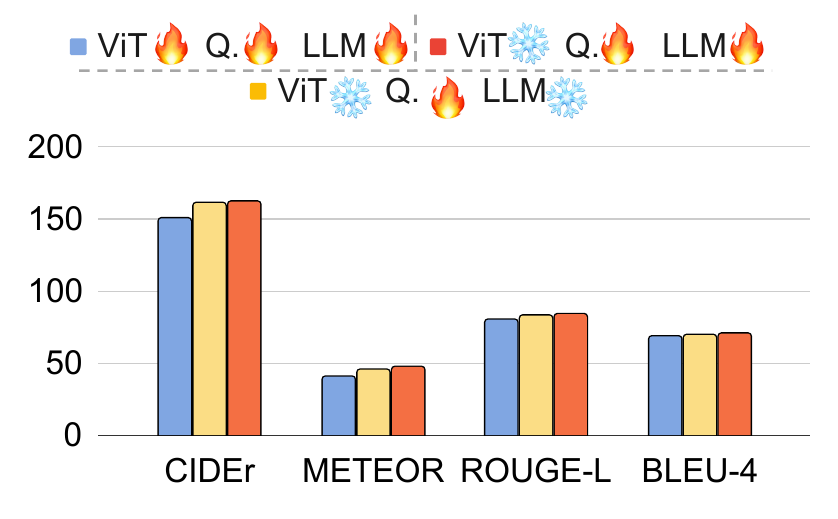}
        \label{fig:trainable-part-msvd}
    \end{subfigure}
    \hfill
    \begin{subfigure}[t]{0.2455\textwidth}
        \centering
        \includegraphics[width=\textwidth]{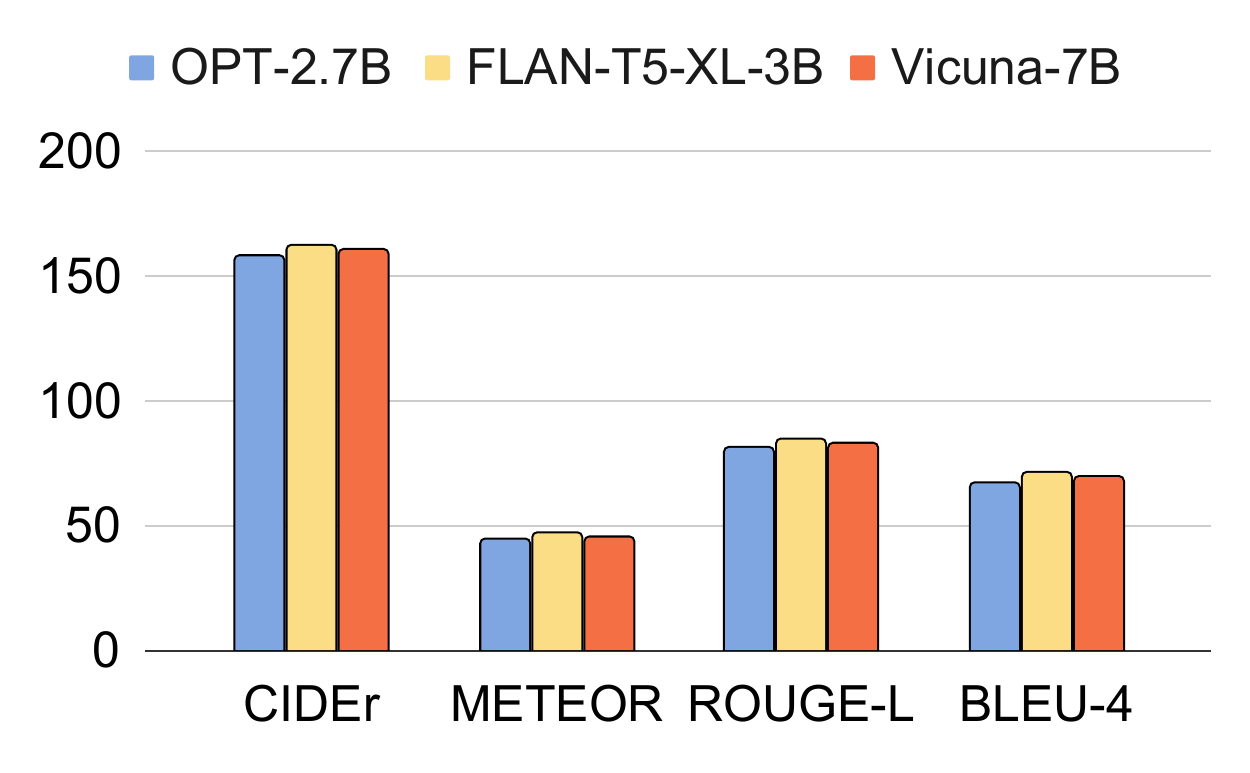}
        \label{fig:small2large-msvd}
    \end{subfigure}
        \hfill
    \begin{subfigure}[t]{0.2455\textwidth}
        \centering
        \includegraphics[width=\textwidth]{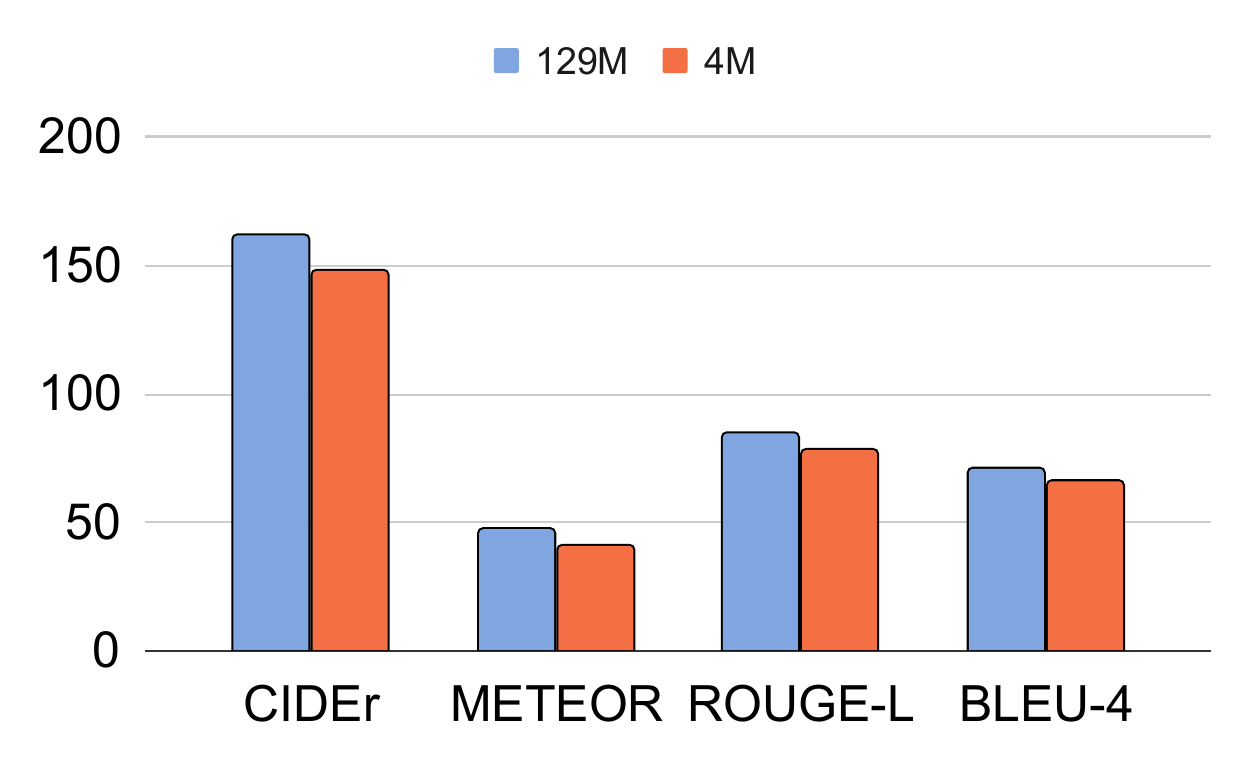}
        \label{fig:pre-train}
    \end{subfigure}
    \hfill
    \begin{subfigure}[t]{0.2455\textwidth}
        \centering
        \includegraphics[width=\textwidth]{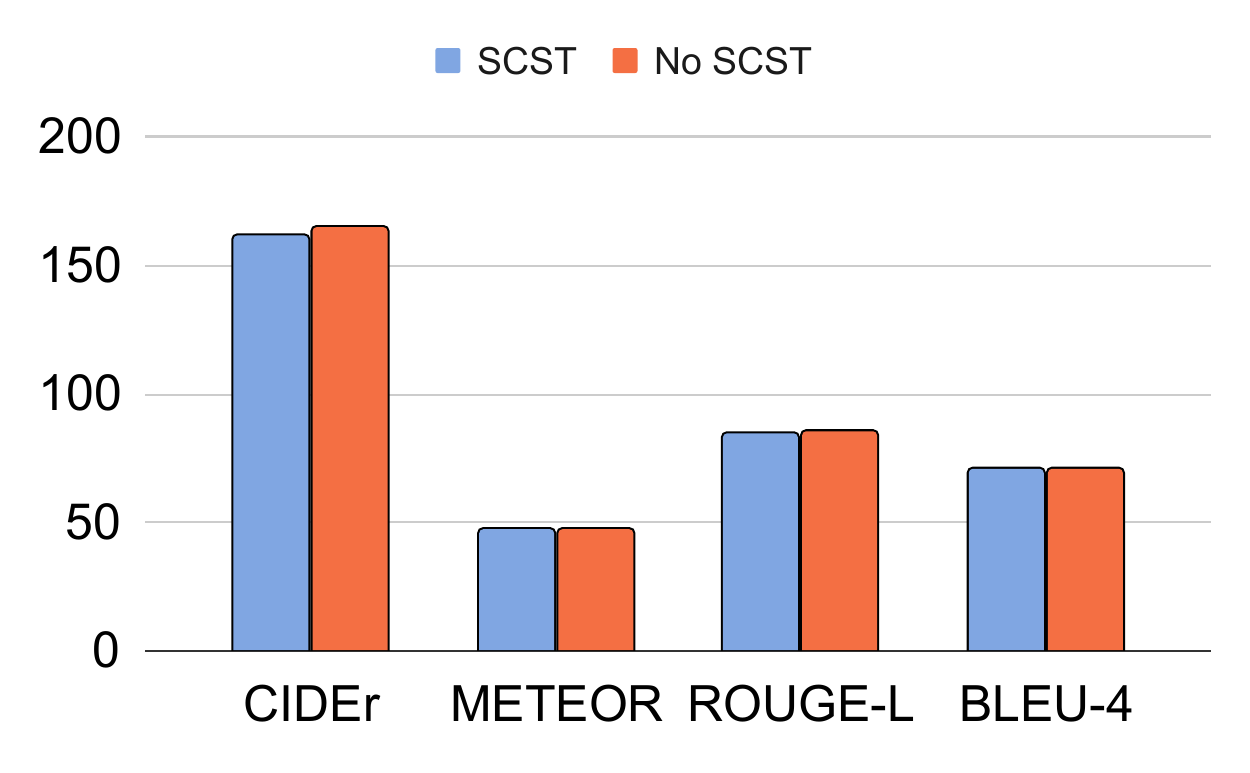}
        \label{fig:scst-bar-msvd}
    \end{subfigure}
    \vspace{-0.2in}
    \caption{Comparative analysis of different training setups for video captioning models on \textit{MSVD} dataset: (a) freezing modules, (b) scales of LLMs, (c) amount of pre-trained image-text pairs, and (d) models trained with and without SCST.}
    \label{fig:msvd}
\end{figure*}

\begin{figure*}[t]
    \centering
    \begin{subfigure}[t]{0.2455\textwidth}
        \centering
        \includegraphics[width=\textwidth]{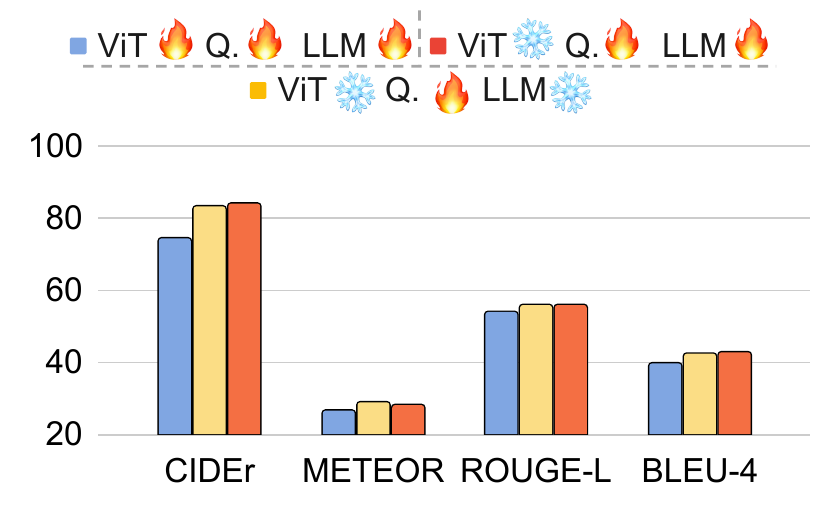}
        \label{fig:trainable-part-msvd-1}
    \end{subfigure}
    \hfill
    \begin{subfigure}[t]{0.2455\textwidth}
        \centering
        \includegraphics[width=\textwidth]{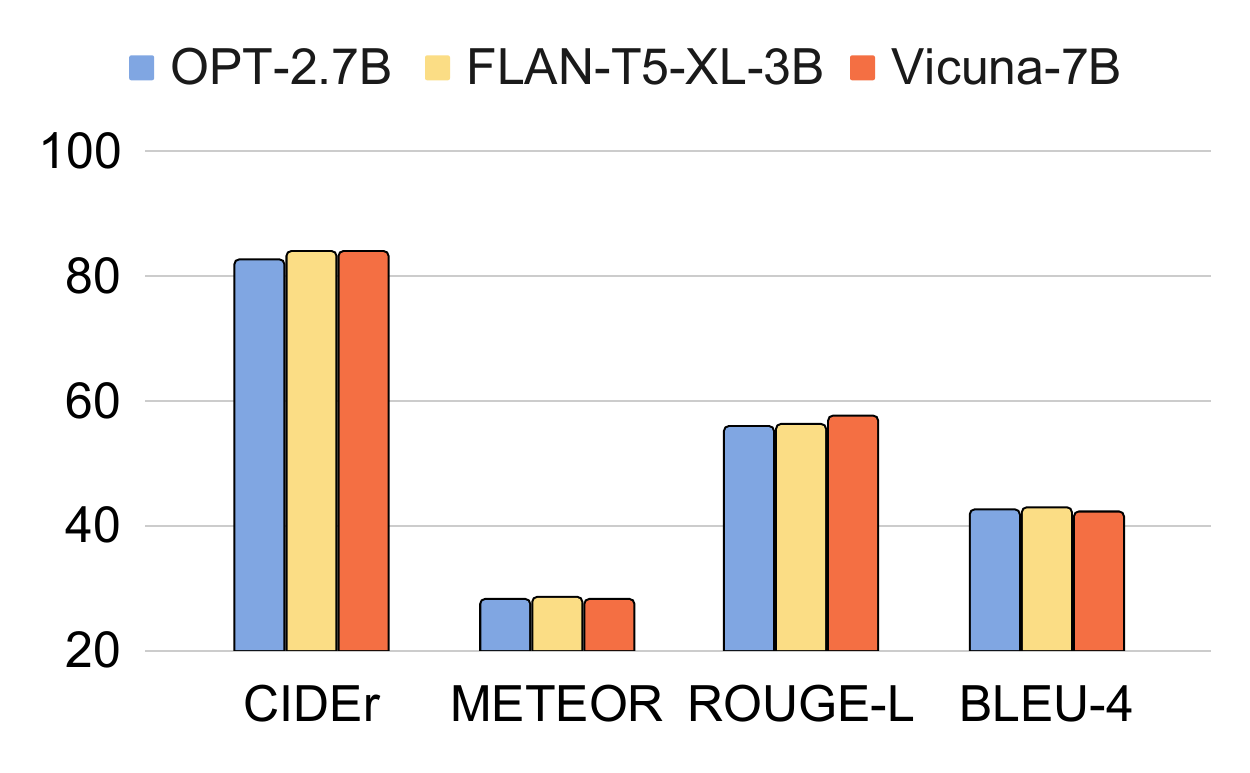}
        \label{fig:small2large-msvd-1}
    \end{subfigure}
        \hfill
    \begin{subfigure}[t]{0.2455\textwidth}
        \centering
        \includegraphics[width=\textwidth]{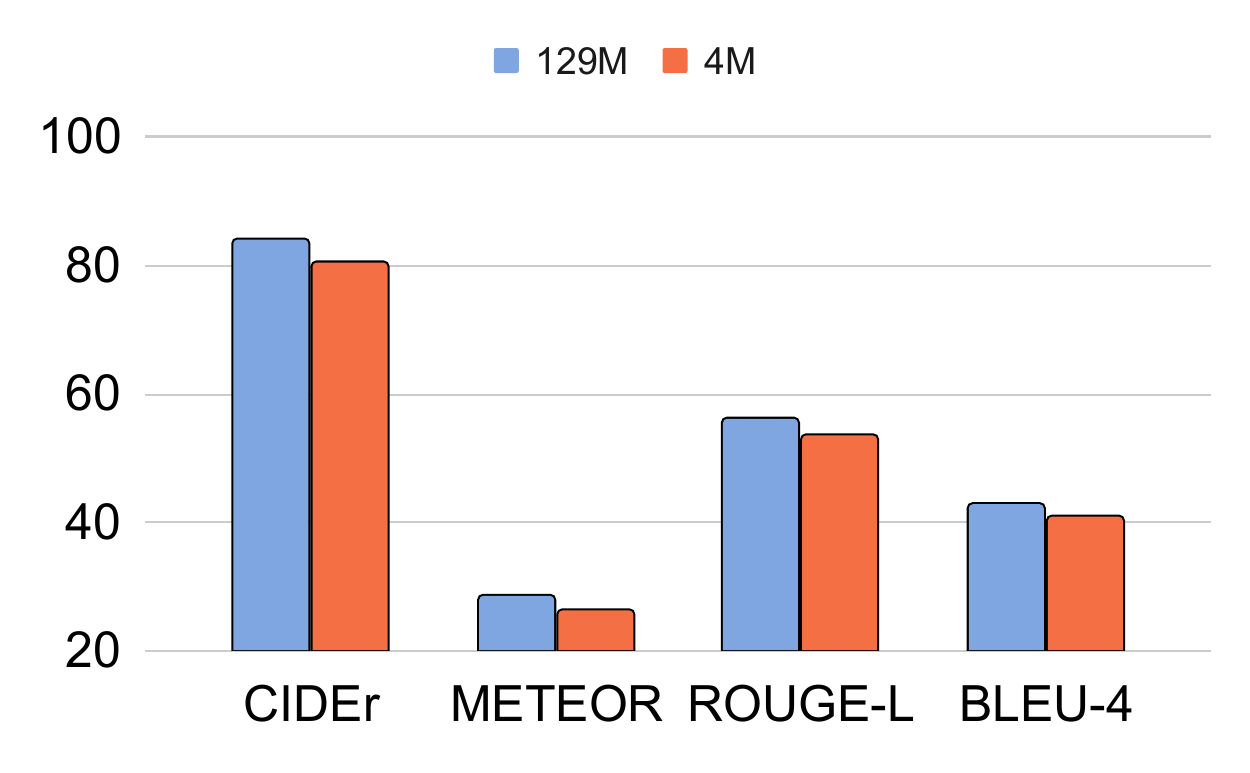}
        \label{fig:pre-train-1}
    \end{subfigure}
    \hfill
    \begin{subfigure}[t]{0.2455\textwidth}
        \centering
        \includegraphics[width=\textwidth]{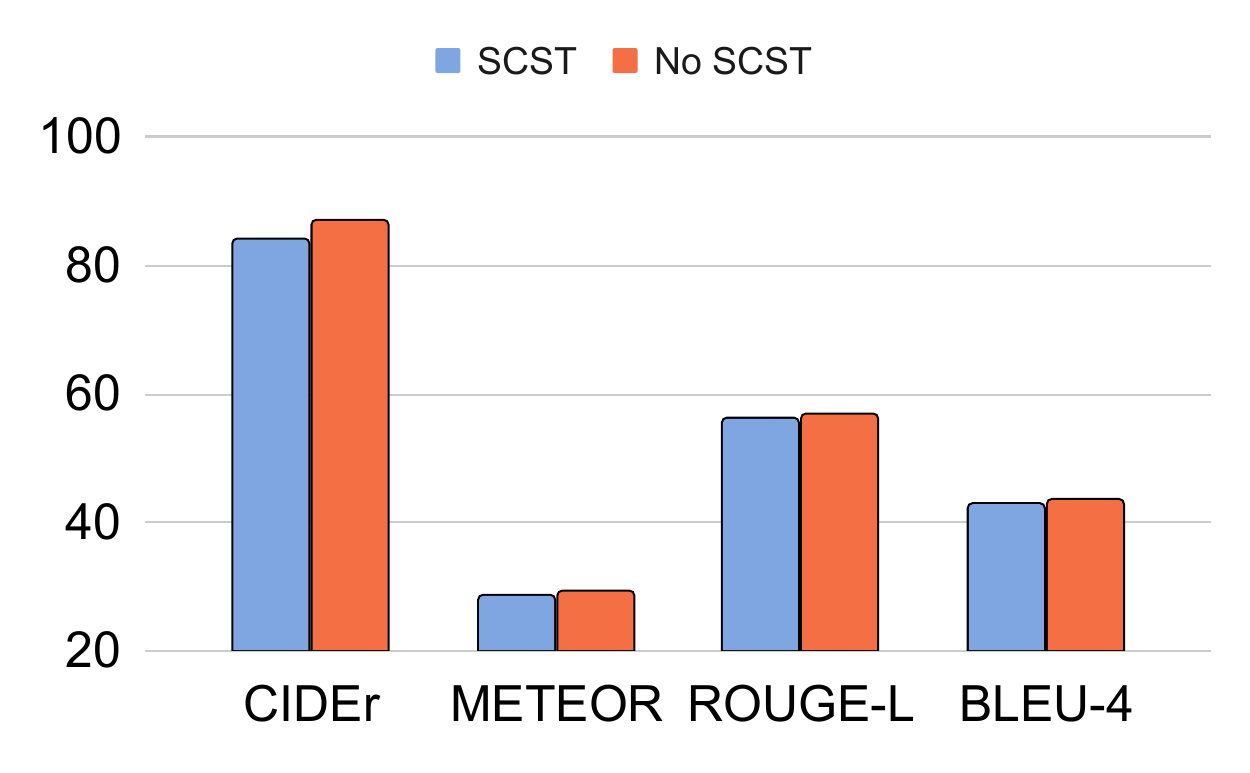}
        \label{fig:scst-bar-msvd-1}
    \end{subfigure}
    \vspace{-0.2in}
    \caption{Comparative analysis of different training setups for video captioning models on \textit{VATEX} dataset: (a) freezing modules, (b) scales of LLMs, (c) amount of pre-trained image-text pairs, and (d) models trained with and without SCST.}
    \label{fig:vatex}
\end{figure*}
The ablation results on the \textit{MSVD} and \textbf{VATEX} dataset are provided in Fig.~\ref{fig:msvd} and \ref{fig:vatex}. The experiments on the \textit{MSVD} and \textbf{VATEX} dataset are primarily aligned with the analysis based on MSR-VTT presented in Sec.~\ref{sec:msvtt}, App.~\ref{sec:app:1}, \ref{sec:app:2}, and \ref{sec:app:3}.

Fig.~\ref{fig:msvd} and \ref{fig:vatex} present detailed comparisons of different training setups for video captioning models on the MSVD and VATEX datasets. We use Fig.~\ref{fig:msvd} as the example, and the results provide the following key patterns across four configurations:
\begin{itemize}
    \item Module freezing (Fig.~\ref{fig:msvd}(a)): The results show that freezing various modules has a significant impact on performance. Models with no frozen components achieve the highest CIDEr scores, indicating the benefit of fine-tuning all parts. However, freezing both LLM and ViT results in the lowest performance, suggesting that the trainability of the connector (Q-Former) and LLM is essential for optimal fitting.
    \item LLM scales (Fig.~\ref{fig:msvd}(b)): Moderate-size LLMs, such as the Flan-T5-XL-3B, provide strong performance across all metrics. Although larger models such as Vicuna-7B offer slight improvements, the gains are modest, likely reflecting MSVD's higher text quality requirements. This finding supports the use of mid-range LLMs as a balanced choice for video captioning tasks.
    \item Pre-training of image-text pairs (Fig.~\ref{fig:msvd}(c)): Models pre-trained on larger datasets (129M image-text pairs) outperform those trained on smaller datasets (4M pairs), especially in terms of CIDEr scores. This result underscores the importance of extensive pre-training for capturing diverse visual-linguistic relationships and improving video captioning performance.
    \item SCST (Fig.~\ref{fig:msvd}(d)): Applying SCST improves the model's ability to generate human-like captions by optimizing directly for the CIDEr metric. Models trained with SCST show noticeable improvements in all evaluation metrics, highlighting its effectiveness in aligning speech generation with human preferences.
\end{itemize}
Overall, the ablation results confirm that flexible tuning of the connector and LLM components is critical for adapting image-text models like BLIP-2 to video captioning tasks. While moderate-sized LLMs offer a balanced trade-off between performance and computational efficiency, extensive pre-training on large datasets significantly improves model performance. In addition, reinforcement learning via SCST effectively improves the quality of generated captions by aligning the training goal with human-centric evaluation metrics.

\subsection{Experiments on MSR-VTT and MSVD Video Question-Answering Datasets}
\label{app:vqa}
\begin{table}[t]
    \centering
    \resizebox{0.495\textwidth}{!}{
    \begin{tabular}{lcc}
    \toprule
    \textbf{Category} & \textbf{MSRVTT-QA} & \textbf{MSVD-QA} \\ \midrule
    \multicolumn{3}{c}{\textit{Module Trainability}} \\ 
    All modules trainable & 18.1 & 36.2 \\ 
    Unfreeze Q-former only & 23.9 & 38.8 \\ 
    Freeze ViT only & 22.5 & 38.5 \\ \midrule
    \multicolumn{3}{c}{\textit{RL to Human Standard}} \\ 
    SCST Disabled & 23.9 & 38.8 \\ 
    SCST Enabled & 24.1 & 41.0 \\ \midrule
    \multicolumn{3}{c}{\textit{Pretrained Image-Text Pairs}} \\ 
    129M & 23.9 & 38.8 \\ 
    4M & 18.8 & 36.2 \\ \midrule
    \multicolumn{3}{c}{\textit{Language Model Size}} \\ 
    OPT-2.7B & 16.5 & 35.7 \\ 
    FLAN-T5-XL-3B & 23.9 & 38.8 \\ 
    Vicuna-7B & 20.2 & 38.5 \\ \bottomrule
    \end{tabular}}
\caption{Top-1 accuracy comparison for different configurations on MSR-VTT and MSVD VQA datasets.}
\label{tab:video-qa}
\end{table}
The experiments on video question-answering (VQA) tasks using the MSR-VTT and MSVD datasets are summarized in Table~\ref{tab:video-qa}. We extend the instruction tuning recipe from LAVIS~\citep{li2022lavis} and InstructBLIP~\citep{instructblip} by 30K steps to test whether our findings from video captioning are applicable to VQA.
The results in Table~\ref{tab:video-qa} show that many of the patterns observed in video captioning extend well to video question answering:
\begin{itemize}
\item Similar to video captioning, keeping more modules trainable leads to better performance. Specifically, models with all components trainable achieve the highest top-1 accuracy, while freezing only the ViT results in lower performance. This underscores the importance of fine-tuning all components for effective adaptation to VQA tasks.
\item Applying SCST slightly improves the model's ability to generate human-like responses by directly optimizing the metrics used in scoring. This is consistent with our findings in video captioning, where SCST helped improve CIDEr scores by aligning model outputs with human preferences.
\item The use of moderately sized LLMs, such as FLAN-T5-XL, achieves strong performance on both datasets. Although larger models, such as Vicuna-7B, provide slight improvements, the gains are modest, suggesting that mid-range LLMs also provide a good balance between accuracy and computational efficiency for VQA.
\item Similar to video captioning, extensive pre-training on large datasets (129M image-text pairs) leads to better performance than on smaller datasets (4M pairs). This reinforces the importance of diverse visual-linguistic pre-training for improving generalization in both video captioning and VQA tasks.
\end{itemize}
\textbf{Overall, our experiments show that the key findings from our video captioning experiments are transferable to video question-answering tasks.} The tuning of trainable Q-formers and LLMs, the reuse of extensive image-text pre-trained BLIP-2, and the use of reinforcement learning all contribute to improving the performance of video-based models across tasks. This transferability suggests that our summarized guidelines provide a basic but general handbook for building effective multimodal models for video captioning and potentially even other extended tasks.

\end{document}